\documentclass[10pt,sigconf]{acmart}

\usepackage{soul}
\usepackage{latexsym}
\usepackage{amsmath}
\usepackage{gensymb}
\usepackage{pifont}
\usepackage{xspace}
\usepackage{svg}
\usepackage{multirow}
\usepackage{xcolor}
\usepackage{algorithm}
\usepackage[noend]{algpseudocode}
\usepackage{algorithmicx}

\usepackage{multicol}
\usepackage{multirow}
\usepackage{amsmath}

\usepackage{amssymb}

\newcommand{\zerodisplayskips}{%
  \setlength{\abovedisplayskip}{2pt}%
  \setlength{\belowdisplayskip}{2pt}%
  \setlength{\abovedisplayshortskip}{2pt}%
  \setlength{\belowdisplayshortskip}{2pt}}
\appto{\normalsize}{\zerodisplayskips}
\appto{\small}{\zerodisplayskips}
\appto{\footnotesize}{\zerodisplayskips}

\newcommand{\sysName}{\mbox{InstMeter}\xspace}

\usepackage{balance}

\usepackage{enumitem}
\setitemize{topsep=2pt,parsep=0pt,partopsep=0pt,leftmargin=12pt}

\usepackage[inline]{./trackchanges}
\usepackage{subcaption}
\usepackage{listings}
\usepackage{xcolor}
\usepackage[dvipsnames]{xcolor}
\usepackage{tcolorbox} 
\usepackage{subcaption}
\usepackage{minted}
\usepackage{caption}
\usepackage{float}
\usepackage{booktabs}

\usepackage[title]{appendix}

\AtBeginDocument{%
  \providecommand\BibTeX{{%
    \normalfont B\kern-0.5em{\scshape i\kern-0.25em b}\kern-0.8em\TeX}}}

\usepackage{balance}

\newcommand{\blue}[1]{{\color{black}#1}}

\begin{document}

\settopmatter{printacmref=false} 
\renewcommand\footnotetextcopyrightpermission[1]{} 
\pagestyle{plain} 
\setcopyright{rightsretained}

\title{InstMeter: An Instruction-Level Method to Predict Energy and Latency of DL Model Inference on MCUs}

\author{Hao Liu}
\email{h.liu-8@tudelft.nl}
\orcid{0009-0009-8424-9703}
\affiliation{
  \institution{Delft University of Technology}
  \country{Delft, The Netherlands}
}

\author{Qing Wang}
\email{qing.wang@tudelft.nl}
\orcid{}
\affiliation{
  \institution{Delft University of Technology}
  \country{Delft, The Netherlands}
}

\author{Marco Zuniga}
\email{m.a.zunigazamalloa@tudelft.nl}
\orcid{}
\affiliation{
  \institution{Delft University of Technology}
  \country{Delft, The Netherlands}
}

\settopmatter{printfolios=false}
\settopmatter{printccs=false}

\begin{abstract}
Deep learning (DL)  models can now run on microcontrollers (MCUs). Through neural architecture search (NAS), we can search DL models that meet the constraints of MCUs. Among various constraints, energy and latency costs of the model inference are critical metrics. To predict them, existing research relies on \blue{\textit{coarse} proxies} such as multiply-accumulations (MACs) and model's input parameters, often resulting in inaccurate predictions or requiring extensive data collection. In this paper, we propose \textbf{InstMeter}, a predictor leveraging \blue{MCUs' \textit{clock cycles}} to accurately estimate the energy and latency of DL models. \blue{Clock cycles} are fundamental metrics reflecting MCU operations, directly determining energy and latency costs. Furthermore, a unique property of our predictor is its strong linearity, allowing it to be simple and accurate. We thoroughly evaluate InstMeter under different scenarios, MCUs, and software settings. Compared with state-of-the-art studies, InstMeter can reduce the energy and latency prediction errors by $3\times$ and $6.5\times$, respectively, while requiring $100\times$ and $10\times$ less training data. In the NAS scenario, InstMeter can fully exploit the energy budget, identifying optimal DL models with higher inference accuracy. We also evaluate InstMeter's generalization performance through various experiments on \blue{three ARM MCUs (Cortex-M4, M7, M33) and one RISC-V-based MCU (ESP32-C3)}, different compilation options (-Os, -O2), \blue{GCC versions (v7.3, v10.3), application scenarios (keyword spotting, image recognition), dynamic voltage and frequency scaling, temperatures (21\textdegree C, 43\textdegree C)}, and software settings (TFLMv2.4, TFLMvCI). 
We will open our source codes and the MCU-specific benchmark datasets. 

\end{abstract}

\maketitle

\vspace{-2mm}
\section{Introduction}
\label{sec_introduction}

\begin{figure}[t]
    \centering
    \vspace{2mm}
    \includegraphics[width=\linewidth]{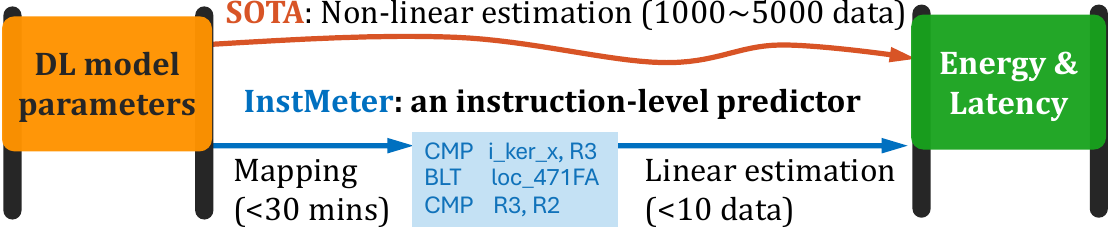}
    \vspace{-6mm}
    \caption{The overview of our method.}
    \label{fig_overview}
    \vspace{-6mm}
\end{figure}

Deep learning (DL) models are increasingly being deployed on microcontrollers (MCUs)\footnote{{MCUs play a pivotal role in empowering a vast array of applications, with global shipments reaching billions annually. MCUs account for roughly 4 out of every 5 processors shipped in the electronics industry today \cite{ESourcing2025EdgeAI}.}}, enabling the development of compact yet intelligent devices. This advancement enables novel systems, such as smartwatches that recognize voice commands and earbuds that autonomously reduce noise~\cite{smartwatch,GreenWaves2025}.
However, designing low-end DL models is not easy. Compared to powerful processors (CPUs, GPUs), MCUs have limited resources but still need to balance accuracy, latency, and energy. For example, some embedded applications must provide real-time responses, requiring \textit{low latency}~\cite{nanodrones,BF_camera_face_detection}; while others operate on platforms with small batteries or even 
without batteries, requiring \textit{low power} consumption~\cite{protean,BF_under_water,BF_face_recognition,solarml}.

To facilitate the design of compact DL models that balance accuracy and efficiency, the community has widely adopted neural architecture search (NAS) methods~\cite{MCUnet, micronets, harvnet, uNAS, Auritus}. NAS methods can evaluate thousands of models to find the most accurate architecture within the required memory, latency, and energy constraints. To achieve this, NAS requires the latency and energy cost of each one of those thousands of models. Since directly measuring latency and energy for every model is time and computationally expensive~\cite{Auritus}, state-of-the-art (SOTA) NAS algorithms rely on \textit{predictors}.

\textbf{The main limitation of the SOTA} is that their energy and latency predictors require thousands of training samples. NAS relies on \textit{model parameters} for estimations~\cite{brp-nas, nnMeter, proxylessnas, micronets, harvnet, uNAS}. The shortcoming is that those parameters have a \textit{non-linear} relation with the final values linked to latency and energy \cite{nnMeter}. Due to that non-linear relationship, SOTA predictors require thousands of samples to capture the intricate relation between model parameters and performance. 

The non-linearity faced by SOTA predictors is due to the optimizations performed on the original model (source code) to obtain the final instructions \blue{(disassembled code, \textit{hereafter abbreviated as disasm})~\cite{nnMeter}}. 
Ultimately, what determines the latency and energy costs of a model are MCU’s \textit{clock cycles}, derived from the executed instructions, each having a fixed cycle cost.
Unfortunately, model parameters only account for architectural variations, e.g., layer types and configurations, without considering lower-level execution details. These factors, such as code implementation details and compiler optimizations, can greatly impact execution efficiency and resource consumption. Thus, they~can introduce latency and energy estimation deviations that require significant training data to obtain an accurate predictor.

{\textbf{To address SOTA limitations}, we propose \textbf{\sysName}, an instruction-level framework that enables \textit{rapid}, \textit{accurate} construction of \textit{linear} 
latency/energy predictors (Figure~\ref{fig_overview}).}
\textit{Contrary to SOTA
requiring
thousands of training samples, \sysName achieves accurate predictions with less than ten training samples, a 100$\times$ improvement.}

\textbf{Challenge.} The design of our predictor requires addressing a critical research challenge in the SOTA: \textit{Overcoming the non-linear relation between model parameters and the model's energy consumption and latency performance.}

A fundamental shortcoming in SOTA is the lack of a clear understanding of the relationship between model parameters and model performance. Current studies see the relationship as a \textit{black-box} that needs to be measured or estimated, and there are three main approaches to do that.
\textit{First}, direct measurements~\cite{Auritus}. This approach is time-consuming and complex, as it requires connecting equipment to perform real measurements. 
\textit{Second}, operation predictors~\cite{harvnet, uNAS}. This method maps model parameters to features that have been proven to be inaccurate, such as multiply-accumulate operations (MACs). 
\textit{Third}, machine learning models~\cite{nnMeter, brp-nas}. This recent group of studies estimates latency using machine learning models. It is more accurate than operation predictors, but requires thousands of training samples. 

These three approaches do not try to understand the non-linearity of the \textit{`black-box'} performance but rather propose methods to estimate it. Thus, a critical research question is to analyze if that non-linear property is inherent to performance estimation or if it can be avoided.

\blue{To address this challenge in our \sysName}, we map model parameters to \blue{MCUs' clock cycles}. 
By doing this, our \sysName requires \textit{only one input}: the total number of \blue{clock cycles} executed by the model, and those \blue{clock cycles} have a natural \textit{linear} relationship with performance metrics such as energy and latency.
\textit{Thanks to the linearity of our model,  we do not need a large amount of training data}. We only need two simple phases. \textit{First}, we build a library mapping model operators to their execution cycles. \textit{Second}, we gather only a handful of training data to build a predictor for our linear model.

\textbf{Contributions.} Our main contributions are as follows. 

\textit{Contribution 1: Transforming the energy and latency estimation of DL models from a non-linear paradigm to a linear one.} We show that DL models can be accurately converted to clock cycles, which have a linear relationship with latency and energy cost. 
When DL models run on MCUs, each model layer is implemented as a set of functions. The compiled instructions of these functions remain nearly identical across different models. The primary variation comes from the values of the \texttt{loop} variables. In other words, when the model parameters change, they affect the number of times that~\texttt{for} or \texttt{while} loops are executed. 
If we can determine these loop values, we can infer the number of \blue{clock} cycles associated with each function to build a linear predictor. 

\textit{Contribution 2: Proposing a novel source-\blue{disasm} code mapping to build accurate and simple predictors.} {Building a linear predictor is not simple because neither the source code nor the \blue{disasm} code has all the required information.} 
The source code (C++) contains the values of the \texttt{loop} variables but no instruction-level details, while the \blue{disasm} code has the opposite trade-off: it has instruction-level information but the variable values cannot be easily inferred. 
We map the \texttt{loops} between these two representations to obtain both instruction-level details and loop execution behavior.
Our mapping, however, does not use DL models (due to their high overhead); we propose a customized algorithm to achieve fine-grained mappings between source and \blue{disasm} code with low time costs. The key benefit of our approach is that we do not need to measure thousands of models to build a training set; we only need a handful of training data.  

\textit{Contribution 3: Creating a dataset to benchmark the performance of DL models running on MCUs.} While the SOTA has some well-known datasets to benchmark the latency of advanced processors (GPU, TPU, CPU)~\cite{nas-bench-201}, there are no equivalent datasets for MCUs. Our first attempt considered using datasets designed for mobile devices (CPUs), such as NAS-Bench201~\cite{nas-bench-201}. However, out of 1000 models randomly chosen from that dataset, only 13 fit the memory constraints of MCUs after compiling them with the TensorFlow Lite for Microcontrollers (TFLM). To evaluate the effectiveness of \sysName, we build a dataset covering three ARM MCUs (Cortex-M4, M7, M33) and \blue{one RISC-V MCU (ESP32-C3)}, two compiler optimizations (-Os, -O2), \blue{GCC versions (v7.3, v10.3), application scenarios (keyword spotting (KWS), image recognition), different dynamic voltage and frequency scaling (DVFS) settings, temperatures (21\textdegree C, 43\textdegree C)} and two TFLM versions (v2.4, vCI). In general, our dataset contains 2125 models with ground-truth measurements for energy and latency considering two applications: KWS and CIFAR10-based image recognition \cite{tinyml}. 
We will open our datasets and source codes.

\textit{Contribution 4: Performing a thorough evaluation to quantify the advantages of a linear predictor.} Our instruction-level predictor requires as few as five training samples to provide competitive performance. Our evaluation shows that, for 90\% of the models, \sysName achieves a relative error below $\sim$30\% with just 5 samples. On the other hand, nn-Meter~\cite{nnMeter} achieves a 90th percentile error of $\sim$100\% with 500 training samples, that is 100$\times$ more training data while providing an error that is 3$\times$ worse. MAC-based predictors~\cite{harvnet, uNAS} have a 90th percentile error of $\sim$200\% across various training sample sizes (5, 50), that is up to 10$\times$ more training data with an error that is 6.5$\times$ worse.
We also test \sysName considering NAS methods. Our predictor identifies optimal models that remain closer to the provided constraints, maintaining an error rate within 10\%, while other predictors exhibit errors up to 100\%. To support further adoption, we provide an instruction dictionary for ARM MCUs Cortex-M4, M7, M33, \blue{and RISC-V MCU ESP32-C3}, allowing researchers to predict instruction types and counts for any DL models.

\section{Background and Motivation}
\label{sec_background}

\begin{figure}[t]
    \centering
    \includegraphics[width=.9\linewidth]{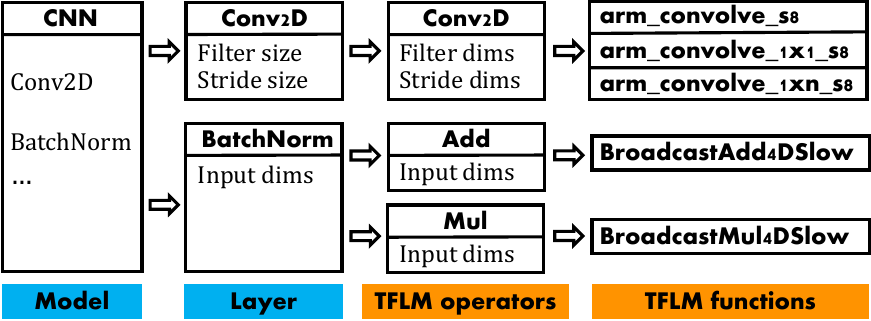}
    \vspace{-1mm}
    \caption{From DL models to TFLM functions.}
    \label{fig_tflm}
    \vspace{-4mm}
\end{figure}

Our \sysName centers around mapping the source code to the \blue{disasm} code, but first, we need to understand $(i)$ how~DL frameworks work on MCUs, $(ii)$ how those frameworks perform code optimizations, and $(iii)$ how SOTA tackles the nonlinear relationship introduced by those MCU frameworks.

\subsection{TFLM Framework}
\label{subsec_tflm_framework}

The TensorFlow Lite for MCUs (TFLM) is a widely employed open-source framework that allows DL models to run efficiently on resource-constrained MCUs~\cite{TFLM}. Other popular frameworks like Edge Impulse~\cite{edgeimpulse} and STM32 AI~\cite{STM32AI}, are built upon it. TFLM consists of \textit{core operators} that execute different parts of a DL model. A single model layer can map to one or more operators. For example, as shown in Figure \ref{fig_tflm}, the \textit{Conv2D} layer corresponds only to the \textit{Conv2D} operator, while the \textit{BatchNormalization} layer requires using the \textit{Add} and \textit{Mul} operators. This modular design enables TFLM to efficiently support a variety of DL models on MCUs.
\blue{
Besides, multiple operators can be fused during the model conversion process (e.g., fused activations), resulting in optimized \texttt{TFLite} models.
TFLM then deploys these optimized model on MCUs, inherently performing these fusion optimizations.}

\vspace{-2mm}
\subsection{TFLM Optimizations}
\label{subsec_tflm_optimization}

TFLM uses multiple techniques to optimize its performance for different MCU platforms and compilers.

\textit{1) Kernel functions.}
TFLM operators are run on kernel functions, but an operator can have multiple kernel implementations; 
\blue{e.g., operator \textit{Conv2D} selects among three functions depending on its kernel dimension:
\texttt{arm\_convolve\_1x1\_s8\_fast} for $1\times 1$ kernels,
\texttt{arm\_convolve\_1\_x\_n\_s8} for $1\times n$ kernels,
and \texttt{arm\_convolve\_s8} for other kernel sizes,
as illustrated in \autoref{fig_tflm}.}
They improve efficiency in specific cases. E.g., \texttt{arm\_convolve\_1x1\_s8\_fast} converts computationally expensive \texttt{1x1} convolutions into more efficient matrix multiplications, reducing memory usage and execution time.

\begin{figure}[t]
    \centering
    \includegraphics[width=0.82\linewidth]{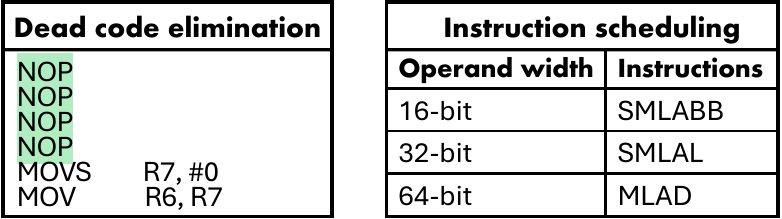}
    \vspace{-1mm}
    \caption{Some optimizations in the TFLM compiler.}
    \label{fig_com_opt}
    \vspace{-4mm}
\end{figure}

\textit{2) Dead code elimination.} Compilers also apply optimizations to enhance TFLM performance on MCUs. A common technique is dead code elimination, which removes unused or unreachable code to reduce program size. Additionally, \verb|NOP| instructions are inserted to ensure instruction alignment and optimize pipeline execution, as shown in Figure~\ref{fig_com_opt} (left). 

\textit{3) Instruction scheduling.} Another common compiling optimization is instruction scheduling. The compiler selects different multiply-accumulate (MAC) instructions based on the operand width—SMLABB for 16-bit, SMLAL for 32-bit, and MLAD for 64-bit operations, as shown in Figure~\ref{fig_com_opt} (right).

All these optimizations enhance performance. However, using one-to-many mappings (kernel functions), adding or removing code (dead code elimination), and selecting different instructions (instruction scheduling) introduces significant non-linear relationships. The last two optimizations are only visible in the compiled \blue{disasm} instructions.

\vspace{-2mm}
\subsection{Limitations of Existing Research}

Existing prediction methods rely on model-level features, which capture only structural information (e.g., model graphs) or high-level operator transformations such as fusion. They ignore instruction-level features, which account for variations in code implementations and compiler optimizations, leading to SOTA requiring excessive training costs.

\textit{$\bullet$ MAC-based methods:} They calculate the number of MAC operations performed by DL models and use regression to correlate the number of MACs with latency \cite{harvnet, uNAS}. They require 1000 training samples and have lower accuracy than other methods because they do not capture non-linear effects.

\textit{$\bullet$ Graph convolutional network (GCN) methods:} They use DL model’s computational graph to estimate performance metrics \cite{brp-nas}. Compared to MAC-based methods, GCNs are effective at capturing nonlinear distortions from code optimizations, but they require around 2000 samples for training.

\textit{$\bullet$ Fusion methods:} They use random forests with multiple layer parameters as features. A notable study~\cite{nnMeter} uses operators as the smallest execution units to fuse different operators (e.g., merging Conv-BN into a single Conv layer). It considers non-linear effects but typically needs 3000 samples.

\textit{$\bullet$ Lookup table (LUT) methods:} They store the energy and latency of fundamental layers \cite{proxylessnas}. Predictions are made by decomposing a model into these layers and summing their stored values. However, building a LUT needs 5000 training points, making it the most time-consuming of all alternatives.

\textbf{Summary.} Current predictors cannot account for source code variations and compiler optimizations. Moreover, among the key SOTA described above, only three tackle MCUs~\cite{micronets,uNAS,harvnet}, the others~\cite{nnMeter,proxylessnas,brp-nas} target powerful processors (CPUs/GPUs/VPUs). 
To address these limitations, we need models that can capture accurately \blue{disasm} instructions and cycle counts since they provide the most direct representation of hardware execution. 
Based on this insight, we propose using the \blue{clock cycles of MCUs} as proxies for estimation. By directly correlating these low-level features with energy and latency, we aim to develop a predictor that is highly accurate and requires little training data (since it is linear).

\section{\sysName Design}
\label{sec_method}

In this section, we present \sysName, an accurate instruction-level linear predictor for the energy consumption ($E$) and latency ($L$) of DL models running on MCUs. By using \textit{a single} instruction-level feature (the number of execution cycles), we improve the estimation accuracy and reduce the amount of training data. The model follows a simple linear equation:
\begin{equation}
    {E} \mathbin{\mathrm{( or }} {L}\mathbin{\mathrm{)}} = a \cdot \text{Cycles} + b,
    \label{eq:main_eq}
\end{equation}
\blue{where $a$ denotes the joules per cycle (for $E$) or 
seconds per cycle (for $L$) and $b$ is a fixed per-inference overhead in joules (for $E$) or seconds (for $L$).} The construction of this predictor has two phases: \textit{instruction profiling} (for the variable $Cycles$) and \textit{data collection} (to determine $a$ and $b$). 
\textit{Instruction profiling}, traditionally a costly step, is significantly optimized in our approach with a novel technique to map source code to \blue{disasm} code. 
After that, \textit{data collection} involves gathering a small number of training samples
to determine the parameters $a$ and $b$, keeping the overall process lightweight.

\vspace{-2mm}
\subsection{Instruction Profiling}
Our goal is to map the DL model code to the \blue{disasm} code to obtain the executed instructions and their corresponding number of execution cycles. This goal could be attained with traditional instruction profiling methods, but those approaches have important disadvantages: \textit{dynamic profiling} requires hardware performance counters or code modifications to track instruction execution \cite{inst_level_pred1, inst_level_pred2, inst_level_pred3,isa-energy,maple,instr}, while \textit{static profiling} uses machine learning models that require hundreds of thousands of training data~\cite{staticPro1, staticPro2, staticPro3}.

We propose a new profiling method that does not require hardware performance counters, code modifications, or a large amount of training data. Our method has the following foundation: \textit{Since all DL models in TFLM are run with a list of specific operators, if we can map the source code of these operators to their corresponding \blue{disasm} code, we can map the source code of any DL model to its corresponding \blue{disasm} code.} Next, we present the insight that allows us to perform low-overhead mapping. Then, we describe our mapping solution.

\vspace{-1mm}
\subsubsection{Theoretical foundation.} Calculating the number of cycles run by a DL model requires two pieces of information $(i)$ the instruction sets and $(ii)$ the number of times these instruction sets are run inside loops. While these two pieces of information are not readily available from either the source code or the \blue{disasm} code, they can be obtained by combining them, as substantiated by our next two observations. 

\begin{figure}[t]
    \centering
    \includegraphics[width=\linewidth]{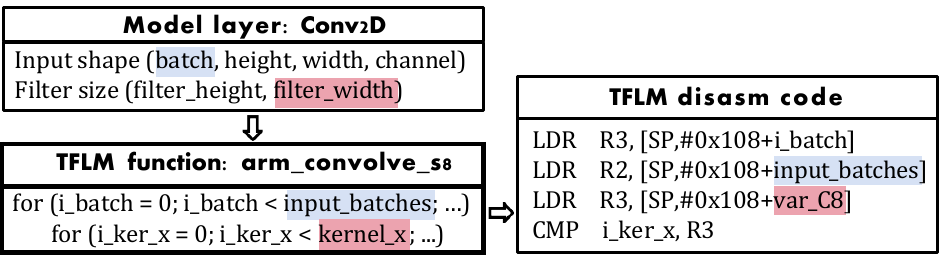}
    \vspace{-4mm}
    \caption{Illustration of the observations.} 
    \label{fig_observation}
    \vspace{-3mm}
\end{figure}

\textit{Observation 1: Within TFLM functions, tuning DL model parameters primarily modifies loop variables (i.e. the number of times a loop is executed), while keeping the code structure unchanged.} On MCUs, DL models are executed~as~a~sequence of TFLM functions. When model parameters are adjusted, the \blue{disasm} instructions associated with these functions remain unchanged; what changes are the variable values, particularly the loop variables. For example, in \autoref{fig_observation}, changing the 
 the parameters such as \textit{batch} size and \textit{filter width} in a \textit{Conv2D} layer only modifies the corresponding loop variables (\texttt{input\_batches} and \texttt{kernel\_x}) in the C++ code (\verb|arm_convolve_s8| function).\footnote{\blue{To further support this observation, we quantify it by comparing kernel functions from two NAS-generated models using Cosine, Jaccard, and LCS similarity metrics. All metrics reached perfect similarity (1.0), confirming that disasm instructions differ only slightly in loop variable values.}} Thus, since the instruction sequence of a function is the same across different models, for any given model, we could obtain the \blue{disasm} instructions associated with the functions required to run the model.

\textit{Observation 2: Loop variable values are difficult to extract from compiled binaries but can be inferred from pre-compiled source code.} Obtaining the \blue{disasm} instructions of each kernel function is necessary but not sufficient. We also need the loop variable values to quantify how many times the \blue{disasm}  code is run. 
Unfortunately, compiled instructions obscure variable names by replacing them with dummy names or memory addresses, making it difficult to extract loop variable values. 
For example, Figure \ref{fig_observation} shows that the compiler renames the loop variable \texttt{kernel\_x} (clearly visible in source code) to a generic identifier (\texttt{var\_C8}), making it harder to trace. 
However, if we are able to map the kernel functions in source code to \blue{disasm} code, we can easily obtain the loop variable values from the source code or model code. 

\vspace{-2mm}
\subsubsection{Our profiling method: A single compilation to profile all kernels} 
The above two observations form the basis of our profiling approach. Based on them, we propose a \textit{loop-level} mapping method that has two key processes: 

\textit{1) Mapping source-\blue{disasm} code:}
First, we extract loop execution counts from the C++ source code. Then, we establish a loop-level mapping to transfer the execution count to the compiled \blue{disasm}. Now, we have the execution count from the source code and instructions \& cycles from \blue{disasm} code.

\textit{2) Estimating the number of cycles:}
We estimate the total execution cycles of the DL model by, first, multiplying each instruction type by the number of times it is executed and by the number of cycles it requires. Then, we sum up the cycles obtained for each instruction type.

\begin{figure}[t]
    \centering
    \includegraphics[width=\linewidth]{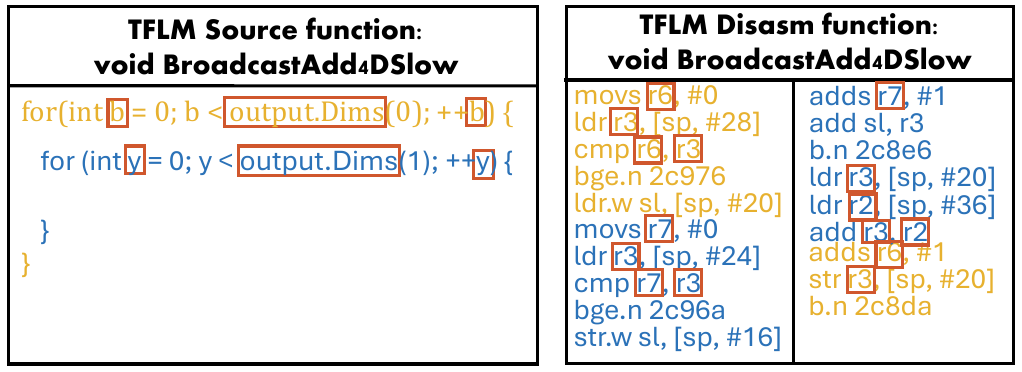}
    \vspace{-4mm}
    \caption{The semantic features of TFLM source and binary functions.}
    \label{fig_lack_fea}
    \vspace{-4mm}
\end{figure}

\subsection{Loop-Level~Source-{disasm} Code Mapping}

Most source-\blue{disasm} code mapping techniques focus on coarse-grained blocks like function-level mapping and library-level mapping, 
which do not fit the fine-granularity of loops~\cite{binpro, b2sfinder,codecmr,cross,binaryai}. Coarse-grained blocks have thousands of lines of code, which contain hundreds of features that allow semantic mapping (such as variable names). On the other hand, loops often have few lines of code with scarce semantic features. 
For example, Figure \ref{fig_lack_fea} shows two loops in source and \blue{disasm} code. In source code, the loops only have one line of code and each of these lines has only two semantic features: (\verb|b|, \verb|output.Dims|) and (\verb|y|, \verb|output.Dims|). In \blue{disasm}, those four variable names are represented with dummy names like \verb|r2|, \verb|r3|, \verb|r6|, and \verb|r7|. This is just a simple example; DL models can have more than one hundred loops (i.e., more than 200 dummy variables), making it hard to perform fine-grained mapping based solely on (obscure) semantic features.

To address this problem, we introduce structural features and a new semantic feature. In brief, our approach consists of two main phases. \textit{First}, we obtain the control flow graphs of the source and \blue{disasm} codes to map their loops based on their graph structures. \textit{Second}, since the graph structures can lead to many-to-many mapping, we exploit semantic~features~to obtain one-to-one loop mapping. Next, we present the details.

\subsection{Structural and Semantic Features} 

\subsubsection{Structural features}

Loops inside code can have different types of relations. The subset ($\subset$) relation describes a loop that is strictly contained within another, and the intersection ($\cap$) relation means two loops that overlap partially. Graph structure is a well-known tool to capture the intersection and subset relationships among loops.  
To obtain structural features, we first generate the control flow graphs (CFGs)~of each source code and then extract the loops within each CFG.

\begin{figure}[t]
    \centering
    \includegraphics[width=\linewidth]{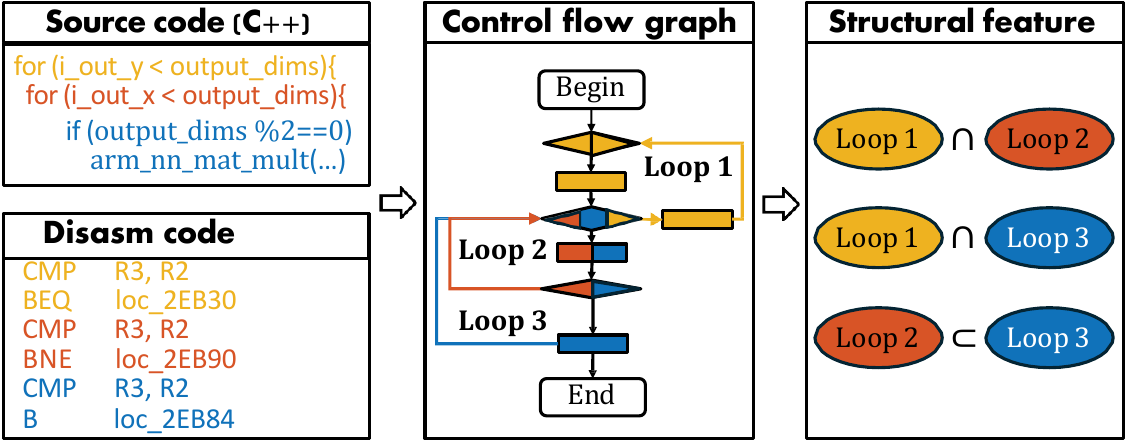}
    \vspace{-4mm}
    \caption{The process of extracting structural features.}
    \label{fig_str_fea}
    \vspace{-4mm}
\end{figure}

\textit{Generating control flow graphs (CFGs).} CFGs contain the code structure and they can be obtained with existing software tools for both the source and \blue{disasm} code \cite{ida_pro, cxx2flow}. Some sample CFGs are shown in \autoref{fig_str_fea}. 

\textit{Extracting loops.} Once the graphs are generated, all the loops in the source and \blue{disasm} graphs can be extracted with various tools, e.g. \textit{Networkx} \cite{networkx}. Figure \ref{fig_str_fea} shows three identified loops using colors to connect them between the CFG and code sources (C++ and \blue{disasm}). For example, \textit{Loop 1} corresponds to the first loop in the source code and \blue{disasm} instructions (yellow color). 

\textit{Obtaining loop relationships.} With the CFGs, we can extract the relationship among loops. By traversing every pair of loops, we can define their relations. If all elements in one loop belong to another loop, we define the subset relation. If partial elements in one loop belong to another loop, we define the intersect relation. For example, in Figure~\ref{fig_str_fea}, \textit{Loop 2} is a subset of \textit{Loop 3} while \textit{Loop 1} intersects with \textit{Loop 2}.

\vspace{-2mm}
\subsubsection{Semantic features.} Semantic features primarily include string literals within loops.

\begin{figure}[t]
    \centering
    \includegraphics[width=1.0\linewidth]{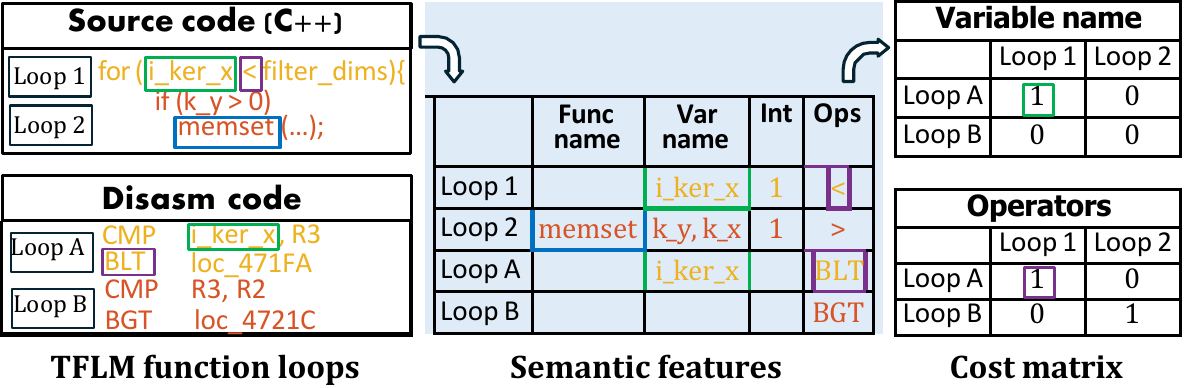}
    \vspace{-5mm}
    \caption{The illustration of semantic features.}
    \label{fig_sem_fea}
    \vspace{-2mm}
\end{figure}

\textit{Traditional semantic features.} String literals such as function names, variable names and integers are widely used in prior studies \cite{b2sfinder, binpro}, but their representations are not consistent. For example,  
as illustrated in Figure \ref{fig_sem_fea}, while the variable name \verb|i_ker_x| appears in both source code (\textit{Loop 1}) and \blue{disasm} code (\textit{Loop A}), the function name \verb|memset| only appears in source code (\textit{Loop 2}), not in \blue{disasm} code. Hence, these traditional features alone cannot be used to map loops. 

\textit{Comparison features.} Unlike previous work, we incorporate comparison operators as semantic features due to their ubiquitous presence: every loop has a comparison operator. For example, in Figure \ref{fig_sem_fea}, while the names of string variables do not remain constant in source and \blue{disasm} code, stop conditions always appear in both. For instance, the conditions \verb|i_ker_x < filter_dims| in \textit{Loop 1} and \verb|k_y > 0| in \textit{Loop 2} are represented with the operators \verb|<| and \verb|>| in source code, and \verb|BLT,BGT| (Branch Less/Great Than) in \blue{disasm} code.

We extract different types of comparison operators and summarize them in \autoref{tab_op_fea}. It is important to note that for each comparison in source code (e.g., \verb|<|), \blue{disasm} instructions use two different representations, e.g., an identical instruction \verb|BLT| (branch less than) and a counter instruction \verb|BGE| (branch greater equal). 

\textit{Generating semantic features.} The semantic features are also obtained from the CFGs. This process is the same as that of generating structural features. 
After extracting the loops, we traverse the source and \blue{disasm} code to record the semantic features of each loop, as shown in \autoref{fig_sem_fea}.

\begin{table}[t]
 \centering
 \caption{The comparison of operator features.}
 \vspace{-2mm}
 \resizebox{\columnwidth}{!}{
 \begin{tabular}{l|c|c|c|c|c}
 \toprule
  \toprule
     Operators (.cpp)& $<$& $\geq$& $>$&$\geq$& $==$ \\ \hline
     Operators (.asm)& \texttt{BLT, BGE} & \texttt{BLE, BGT}  & \texttt{BGT, BLE}  & \texttt{BGE, BLT}  & \texttt{BEQ} \\ \hline 
     Operators (.cpp)& $!=$ & $\gg$& $\ll$ & $\&$ & $|$ \\ \hline
     Operators (.asm)& \texttt{BNE}  & \texttt{ASR, ASRS}& \texttt{LSL, LSLS} & \texttt{AND}& \texttt{ORR}\\ \hline 
 \end{tabular}}
 \label{tab_op_fea}
\end{table}

\subsection{Structural and Semantic Mapping}

Our algorithm maps the loops between source and \blue{disasm} code using the structural and semantic features described in the prior subsection.
We propose a progressive mapping algorithm. We first map structural features based on graph matching algorithms, and then apply commonly used semantic mapping algorithms for the semantic features \cite{b2sfinder}.

\subsubsection{Structural mapping.} 
\begin{figure}[t]
    \centering
    \includegraphics[width=.99\linewidth]{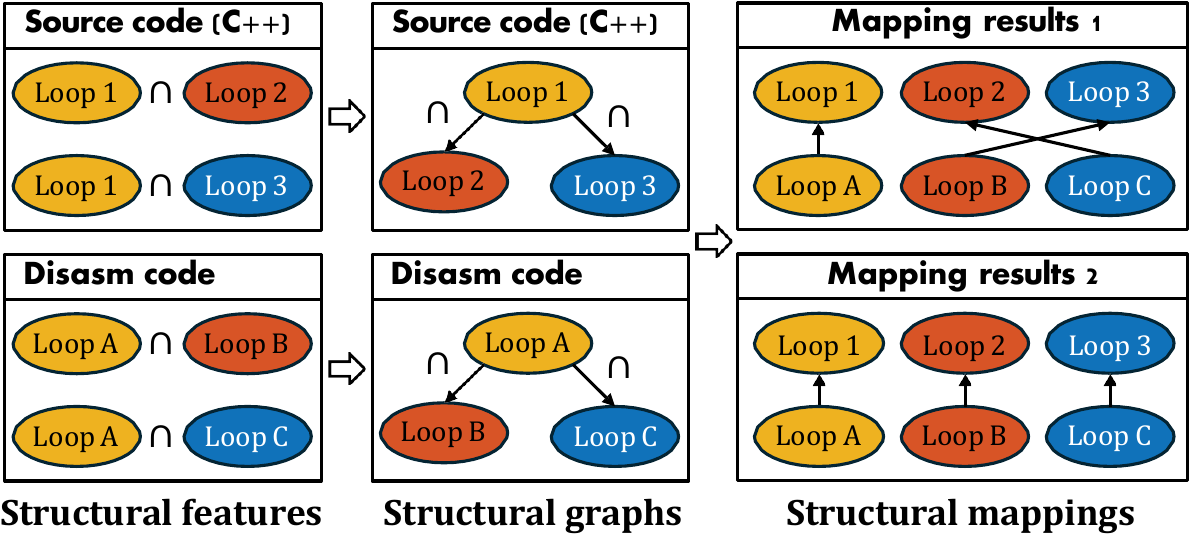}
    \vspace{-1mm}
    \caption{The process of structural mappings.}
    \label{fig_str_map}
    \vspace{-2mm}
\end{figure}

Structural mapping identifies loop correspondences between source and \blue{disasm} CFGs using graph matching. We use the matching algorithm VF2 \cite{vf2} because it is designed for large graphs and uses an efficient search strategy, based on depth-first search \cite{d2s}.
In our case, the graph nodes represent the loops and the edges represent the structural features, as shown in Figure \ref{fig_str_map}. With these graphs, the graph-matching algorithm traverses nodes and edges to identify the most similar pairs. For instance, as shown in Figure \ref{fig_str_map}, \textit{Loop 1} and \textit{Loop A} form a unique match due to their shared characteristic of having exactly two interactions, with \textit{Loops 2 and 3} and \textit{Loops B and C}, respectively. 
However, the mapping is not always unique. For instance, \textit{Loops 2 and 3} and \textit{Loops B and C} exhibit identical structural features, preventing a clear mapping based solely on structural properties. Consequently, the algorithm yields two possible matchings: (Loops 1$\mapsto$A, 2$\mapsto$B, 3$\mapsto$C) and (Loops 1$\mapsto$A, 2$\mapsto$C, 3$\mapsto$B). To overcome this shortcoming, the next phase of our algorithm uses semantic matching.

\vspace{-2mm}
\subsubsection{Semantic mapping.}
When structural features lead~to~inconclusive mappings, we use semantic features to identify the best candidate. Semantic mapping has three steps: building similarity matrix for semantic features, assigning different weights to features, and computing candidate scores.

\textit{Constructing the similarity matrix.} 
Similarity matrices measure the correlation of features between source and \blue{disasm} loops, and we adopt the approach from \cite{b2sfinder} to build our matrices. Some of the undefined relations arising from structural mapping can be resolved with traditional semantic features (e.g., variable names) but others need comparator features. 
E.g., Figure \ref{fig_sem_fea} shows that the similarity score between \textit{Loop 1} (source) and \textit{Loop A} (\blue{disasm}) is high since the \textit{variable name} \verb|i_ker_x| is in both loops.
However, this traditional feature does not work for the \texttt{if} statement, since the function name \texttt{memset} appears in \textit{Loop 2} but not in \textit{Loop B}. In this case, the comparator operator allows us to perform the mapping. 

\textit{Assigning weights to features.} We have four semantic features: \textit{function name}, \textit{variable name}, \textit{integer}, and \textit{comparison operator}. Each feature requires a matrix and provides a varying degree of similarity between source and \blue{disasm} loops. Thus, two important questions arise: \textit{What features are the most important? How much weight should we assign to each feature?} When the relevance of features cannot be quantified clearly, the best way to assign weights is random scalarization \cite{randomScalarization}, which explores diverse weighting combinations to ensure unbiased optimization. The random scalarization and the computation of similarity scores are described next.

\textit{Computing similarity scores.} 
Formally, considering $n$ loops in the source code and $m$ loops in the \blue{disasm} code that have a many-to-many mapping due to the limitations of structural features, let us define four similarity matrices $\mathcal{M}_{f}$, one matrix for each feature $f$, where $s_{f}(i,j)$ is a matrix element capturing the similarity between loop $i$ and loop $j$ for feature $f$.
Considering that we are using random scalarization, let us define four weights $w_i$, where $\sum w_i = 1$, and run 100 random weight vectors to avoid biases. 
Considering the above notation, the total similarity $S(i,j)$ between loops $i$ and $j$ is given by summing the similarities of all features:
$
S(i,j) = \sum_{r=1}^{100} \sum_{f=1}^4 s_f(i,j) w^r_f .
$
The best mapping candidate is the one that maximizes the similarity, that is:
$
S^* = \max_{(i,j) \in (n,m)} S(i,j) .
$

\textit{Algorithm overview.}
The complete procedure is outlined in Algorithm \ref{alg_progress_algo}, which progressively integrates structural and semantic mapping.
For structural matching, we first construct a relation graph (lines 2-3) and apply the VF2 algorithm for graph matching (lines 4-5).
For semantic matching, we generate a similarity matrix for each feature type (lines 8-11) and assign different weights to these matrices (lines 14-15). Finally, we compute the scores for all candidates and select the best match (lines 16-20).

\begin{figure}[t]
\begin{minipage}{\linewidth}
\vspace{-2mm}
\begin{algorithm}[H]
\small
\captionof{algorithm}{Structural and semantic matching algorithms}
\label{alg_progress_algo}
\begin{algorithmic}[1]
\Function{Structural\_Matching}{$structural\_features$}
    \State Construct graphs $G_s = (V_s, E_s)$ and $G_b = (V_b, E_b)$  
    \State Add edges based on $structural\_features$
    \State Initialize mapping: $initial\_map \gets \{0: 0\}$
    \State $matches \gets$ \Call{VF2\_Search}{$G_s$, $G_b$, $initial\_map$}
    \State \Return $matches$
\EndFunction

\Function{Semantic\_Matching}{$matches$, $semantic\_features$}
    \State Initialize $sim\_matrices \gets []$
    \For{each $feature$ in $semantic\_features$}
        \State $sim\_matrix \gets$ \Call{Similarity}{$feature$}
        \State Append $sim\_matrix$ to $sim\_matrices$
    \EndFor

    \State $best\_score \gets -\infty$, $best\_match \gets None$
    \While{multiple top matches exist}
        \State Generate $weights \gets$ \Call{Random\_Scalarization}{}
        \State Compute weighted scores for each match:
        \For{each $match$ in $matches$}
            \State $score \gets$ \Call{Weighted\_Sum}{$weights$, $sim\_matrices$}
        \EndFor
        \State Select highest-scoring match:  
        \State $best\_score \gets$ \Call{Max}{$scores$}
        \State $best\_match \gets$ \Call{Select\_Unique}{$scores$, $best\_score$}
    \EndWhile
    \Return $best\_match$
\EndFunction
\end{algorithmic}
\end{algorithm}
\end{minipage}
\end{figure}

\subsection{Estimate Total Cycles and Generate Instruction Library}
 
\blue{The disasm-source mapping described above needs to be performed only once if the following two conditions are met: (1) the MCU architecture remains unchanged (e.g., Cortex-M4), and (2) neither the source code nor the corresponding disassembled instructions undergo significant changes, such as the introduction of new operator files (validated in Section \ref{subsec_eval_compatibility}).} After that, we build an instruction library that can be used to estimate the total execution cycles of any DL model. First, we will explain how we get the number of cycles per kernel function, and then, describe the instruction library.

\textit{Estimating the number of cycles.}  
For each kernel function $k$, 
total execution cycles are computed by multiplying the occurrence ($o_i$, \blue{via our looping algorithm}) of each instruction type $i$ by \blue{its cycle-per-instruction (CPI, denoted as $c_i$)}:
\begin{equation}
    \text{Cycles}_k = \sum_{i=1}^{N_k} \left(o^k_i \cdot c^k_i\right) ,
\end{equation}
where $N_k$ is the number of instruction types in function~$k$. \blue{The accurate CPI values can be derived from an MCU's official documentation, or from empirical measurements.}\footnote{\blue{
For Cortex-M4, CPI values are directly obtained from the official ARM documentation~\cite{cortexm_manual}.
Due to incomplete official CPI documentation for Cortex-M7 and Cortex-M33, we adopt CPI values from Cortex-M4 as proxies because these MCUs share common instruction types for TFLM inference. Although we did not explicitly validate CPI accuracy, our evaluation (cf. Figure \ref{fig_m7m33_energy}) indirectly supports this approximation.
For the significantly different MCU ESP32-C3, which has a \textit{RISC-V-based architecture}, we empirically measure CPI values by embedding assembly instructions within C code, executing each instruction type on the MCU 100K times, and averaging the results.}}

\begin{figure*}[t]
    \centering
    \includegraphics[width=\linewidth]{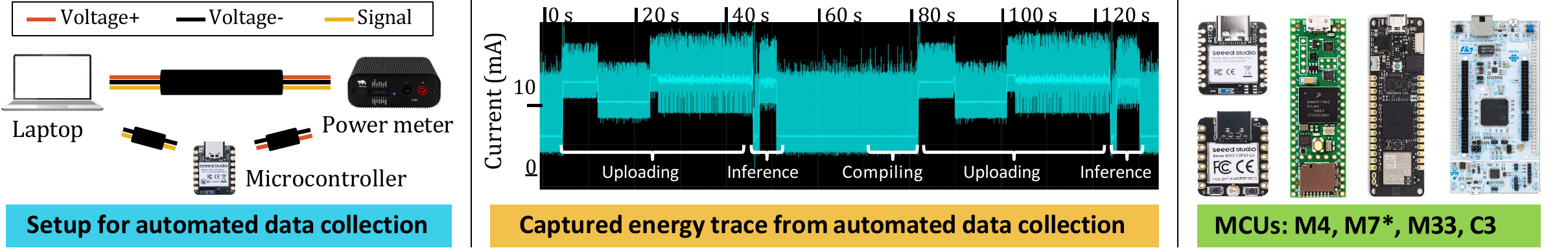}
    \vspace{-3mm}
    \caption{Our tool for automatic energy and latency measurements; captured energy traces are illustrated. MCUs: M4 (Seeed nRF52840), M7 (Teensy 4.1 and STM32F767), M33 (Arduino Portenta C33) and C3 (Seeed ESP32C3).}
    \label{fig_measure_setup}
    \vspace{-2mm}
\end{figure*}

\textit{Generating Instruction Library.} \blue{Once we obtain the number of cycles per function, we calculate the cycles in each operator using the operator-to-function mappings (cf. Section~\ref{subsec_tflm_optimization}).} Then, we build a library to estimate the total number of cycles of any DL models. 
\blue{When a new DL model is provided, we do not need to compile it.} \blue{ We first convert it into a TFLite model, inherently including operator fusions (cf. Section~\ref{subsec_tflm_framework}).} 
Then we use the instruction library to parse the model into the operators and obtain their cycles.
The sum of these cycles is the total cycles executed by the DL model. 
This process enables fast and precise estimation of execution cycles.

To estimate the energy consumption and latency, we only require a few training samples. For these samples (DL models), we measure the energy and latency; and derive their number of cycles using our instruction library. The ground truth and cycles allow us to obtain the parameters $a$ and $b$ for energy and latency in ~\autoref{eq:main_eq}.

To optimize cycle profiling, we pre-generate instruction libraries for common MCUs (Cortex-M4, M7, M33 and ESP32C3). When constructing energy and latency predictors for those MCUs, we can directly use the pre-generated libraries. 
\blue{ The library remains valid unless the kernel source code significantly changes (e.g., new operators added) or a new MCU architecture is introduced. In such cases, users need to provide updated source code, its disassembled instructions, loop execution counts extracted from the source, and MCU-specific CPI values (obtained either from documentation or empirical measurement).} The whole process roughly takes 30 minutes.

\section{Implementation \& Benchmark Datasets}
\label{sec_implementation}

\subsection{ Implementation}
\label{subsec_implementation}

To profile the instructions/cycles for \sysName, we need to analyze the {source} and \blue{disasm} code of TFLM. We build a \textit{single} DL model using 10 commonly used operators in MCUs:
\textit{Conv2D, DepthConv2D, FullyConnected, MaxPool2D, AvgPool2D, ReLU, Add, Mul, Softmax, and Reshape}. These operators cover most cases in MCU NAS \cite{harvnet, uNAS, micronets, MCUnet}. From the source code (C++), we identify that these operators are implemented with 26 kernel functions in TFLM\footnote{The 26 kernel functions include \texttt{arm\_elementwise\_mul\_s8},
\texttt{arm\_element\\wise\_add\_s8}, \texttt{arm\_depthwise\_conv\_s8}, \texttt{arm\_softmax\_s8}, \texttt{arm\_convolve\\\_s8}, \texttt{arm\_max\_pool\_s8}, \texttt{arm\_depthwise\_conv\_8s\_opt}, \texttt{arm\_depthwise\_c\\onv\_3x3}, \texttt{reshapeOutput}, \texttt{ReluQuantized}, 
\texttt{FlatSize}, among others.}. 
For the source code, we identify these kernel functions through their function names. 
For the compiled {\blue{disasm} code}, we extract the 26 kernel functions using IDA Pro~\cite{ida_pro}, a decompilation tool that allows locating functions accurately. 
After the kernel functions are identified in the source and \blue{disasm} code, we obtain the control flow graphs (CFG) using
IDA Pro and cxx2flow~\cite{cxx2flow}.
Once the CFGs are obtained, we use our framework, in Section~\ref{sec_method}, to achieve robust source-\blue{disasm} mapping.

\emph{Automated data collection.}
To evaluate the performance of \sysName, we also design and implement an automatic data collection tool. While similar automation tools have been mentioned in~\cite{Auritus}, they do not describe implementation details. \blue{The setup of our tool is shown in Figure~\ref{fig_measure_setup}. It fully automates model compilation, uploading, execution, and measurement through customized Python scripts. To accurately identify inference intervals from energy traces, we insert short delays before and after each inference, clearly distinguishing inference intervals from idle periods based on power-level differences. Each inference is consecutively executed 10 times to ensure stable, precise energy measurements, and the average consumption is reported using an OTII ACE PRO power meter with nanowatt-level resolution.}

\vspace{-3mm}
\subsection{Creating Benchmark Datasets for MCU-Specific Predictor Evaluation}

The popular NAS-Bench201 dataset~\cite{nas-bench-201} aims at mobile-level devices (CPUs) and is not well suited for MCUs due to two key limitations. First, its models exceed typical MCU memory constraints—we found that only 13 out of 1000 randomly sampled models fit within 256\,KB, a common limit of MCU. Second, the dataset’s model granularity does not align with mainstream MCU design. While NAS-Bench201 adopts a coarse-grained approach with layer-stacked cells, MCUs typically use individual layers rather than predefined cells. 

To bridge this gap, we create two custom datasets measuring latency and energy. The first dataset focuses on testing a diverse set of models, while the second dataset focuses on testing different hardware platforms, software versions, \blue{compiler versions, temperatures, applications, DVFS configurations,} and compiler optimizations. 

\textbf{Performance evaluation dataset.}  
This dataset \textit{consists of two sub-datasets: a mixed-layer dataset ({ML-Data}) and a single-layer dataset ({SL-Data}).} 
\textbf{ML-Data} contains models constructed by randomly mixing layers, allowing us to evaluate the predictor’s accuracy under diverse and heterogeneous architectures. The models are generated using an MCU-based NAS algorithm~\cite{uNAS}, which explores a fine-grained search space optimized for MCU constraints. To ensure consistency, we use Cortex-M4 as the target MCU, compiled with -O0 optimization level, and execute the models within TFLMv2.4. A total of 445 valid models are collected. 
\textbf{SL-Data}, inspired by~\cite{nnMeter}, consists of models only with one type of layer, such as Conv2D or MaxPooling2D, to evaluate the predictor’s accuracy on a layer basis. We collect models for eight layer types, including Conv2D, Conv1x1, DepthwiseConv2D, Dense, MaxPooling2D, and AvgPooling2D. For each layer type, we collect ten samples, leading to a total of 60 models, using the same setup as ML-Data (MCU, optimization level and TFLM version).

\textbf{Compatibility evaluation dataset.}  This dataset evaluates predictor adaptability across diverse hardware, software, and compiler settings, comprising seven distinct sub-datasets: \textbf{MCU-Data}, \textbf{Compiler-Data}, \textbf{\blue{GCC-Data}}, \textbf{\blue{Temp-Data}}, \textbf{\blue{App-Data}}, \textbf{\blue{DVFS-Data}}, and \textbf{TFLM-Data}. 
\textbf{MCU-Data} evaluates performance across four MCU architectures (Cortex-M4, M7, M33, ESP32C3) with fixed compiler optimization (-O0) and TFLMv2.4, collecting 100 samples per MCU (400 samples in total). 
\textbf{Compiler-Data} analyzes sensitivity to compiler optimizations (-Os, -O2) using Cortex-M7 and TFLM v2.4, collecting 100 samples each (200 samples in total). \blue{\textbf{DVFS-Data} assesses robustness under two frequency-voltage pairs (120MHz/1.1V, 216MHz/1.3V) on STM32767ZI MCU, with 100 samples per setting (200 samples in total). \textbf{GCC-Data} considers two compiler versions (GCC v7.3, v10.3) at optimization -O0 on Cortex-M4, collecting 100 samples each (200 samples in total).
\textbf{Temp-Data} evaluates predictor stability at two MCU temperatures (21\textdegree C, 43\textdegree C) using Cortex-M4, fixed compiler (-O0) and TFLMv2.4, collecting 100 samples per temperature (200 samples in total). \textbf{App-Data} tests generalization across keyword spotting and image classification (CIFAR10), each scenario collecting 100 samples (200 samples in total), using Cortex-M4, compiler -O0, and TFLMv2.4. }
Finally, \textbf{TFLM-Data} examines software version impact (TFLMv2.4, TFLMvCI) on Cortex-M4 with -Os optimization, collecting 100 samples per version (200 samples in total).

\textbf{Summary of the data collection.}  
Across all the datasets, we collect a total of 2125 data points. \blue{Except for the \textbf{SL-Data}, models are generated using an MCU-based NAS algorithm \cite{uNAS}, which explores a fine-grained search space optimized for MCU constraints.}
The time required to compile and collect each sample ranged from 1 minute (on Cortex-M7, M33, ESP32C3) to 5 minutes (on Cortex-M4), leading to a total data collection time exceeding 5000 minutes. We collect these comprehensive datasets not only for our evaluation but also hope that they are valuable for our community to evaluate DL models' performance on MCUs.

\vspace{-2mm}
\section{Performance Evaluation}
\label{sec_eval}

{\textbf{Baselines.}} 
We compare our \sysName with two widely used state-of-the-art (SOTA) methods. \emph{$(1)$MACs-based predictors} \cite{uNAS, harvnet}, which estimate performance based on the number of MAC operations in a model and apply regression methods for prediction. \emph{$(2)$nn-Meter} \cite{nnMeter}, which defines fused kernels (layers) and utilizes a random forest approach combined with adaptive sampling methods to make predictions.

{\textbf{Evaluation metrics.}} 
We use the \textit{relative error rate} as the main evaluation metric, defined as $\frac{|y_\text{pred} - y_\text{true}|}{y_\text{true}} \times 100\%,$ where $y_\text{pred}$ is the predicted value and $y_\text{true}$ is the ground truth. Unlike other metrics such as $R^2$ or mean squared error (MSE), which can be disproportionately affected by outliers or large values, the relative error rate measures the proportional difference between predicted and true values, ensuring a fair and interpretable evaluation across all data points.

\begin{figure}[!t]
    \begin{minipage}[t]{\columnwidth}
        \centering
        \begin{subfigure}{\columnwidth}
        \centering
        \includegraphics[width=\columnwidth]{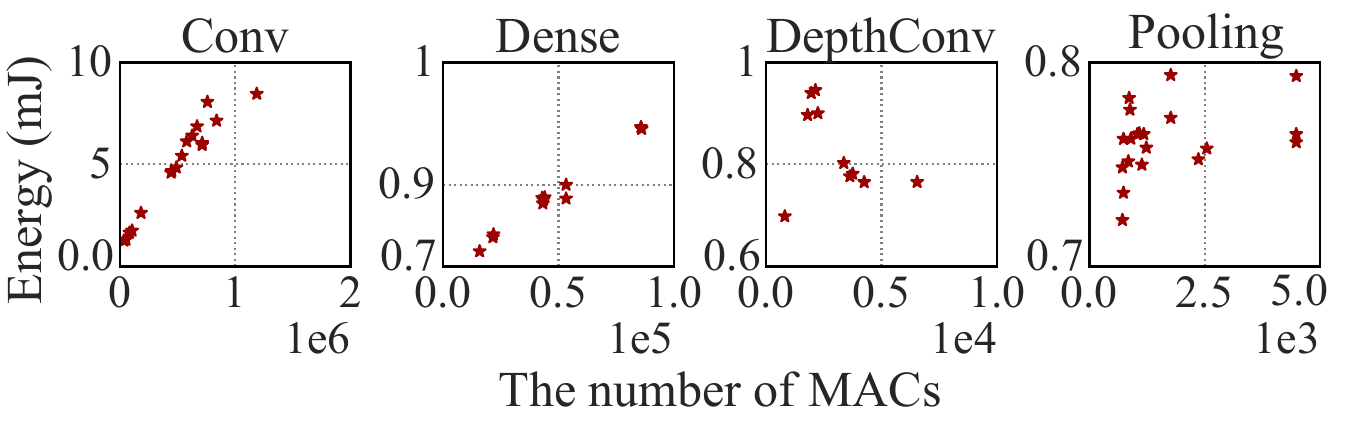}
        \vspace{-6mm}
        \caption{MACs-based predictor: Energy vs number of \textit{MACs}}
        \end{subfigure}
        \begin{subfigure}{\columnwidth}
        \centering
        \includegraphics[width=\columnwidth]{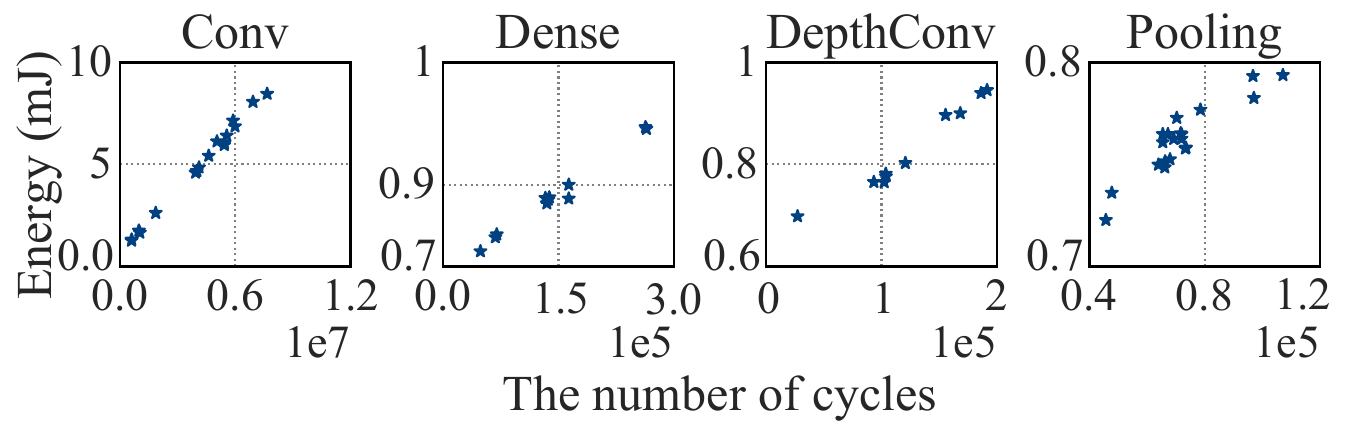}
        \vspace*{-6mm}
        \caption{Our cycle-based \sysName: Energy vs number of \textit{cycles}}
        \end{subfigure}
        \vspace{-7mm}
        \caption{Linearity between our proxy and measured \textit{energy} on pure-layer models.\protect \footnotemark \textit{\textmd{(Results on other layers, such as Residue, are similar and omitted due to space limitations.)}}}
        \label{fig_MAC_inst}
        \vspace{-1mm}   
    \end{minipage}
    \vspace{-1mm}
\end{figure}

\footnotetext{Note that this figure does not directly represent the energy cost of each layer in DL models; the actual cost must also take into account the number of layers in the model and the input parameters.} 

{\textbf{Training methods.}}{Unless otherwise stated, we only use 5 training samples~for our InstMeter. {Given the data scarcity, we employ a sub-sampling strategy: we iterate through internal training/validation splits (e.g., train on 2, validate on 3) to select the regression line that minimizes internal error. This optimal model is then evaluated on the global test set. We repeat this process 10 times with independent random seeds to ensure statistical robustness.}
\textit{Due to space limitation, from Section~\ref{subsec_sota} we only present the results on energy prediction; \textbf{the results on latency prediction are similar and provided in Appendix.}}}

\begin{figure}[t]
        \begin{subfigure}{\columnwidth}
        \centering
        \includegraphics[width=\columnwidth]{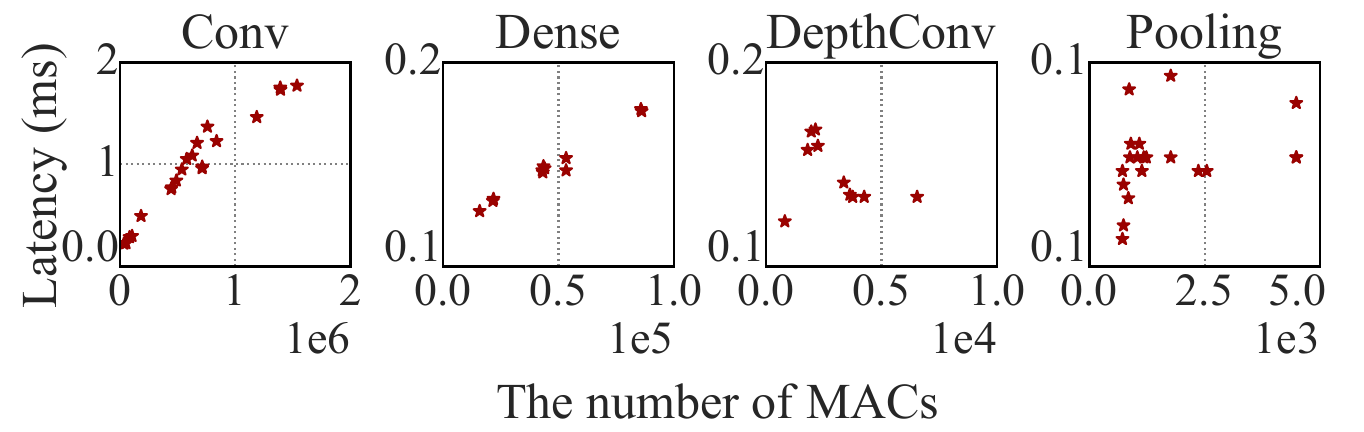}
        \vspace{-6mm}
        \caption{MACs-based predictor: Latency vs number of \textit{MACs}}
        \end{subfigure}
        \begin{subfigure}{\columnwidth}
        \centering
        \includegraphics[width=\columnwidth]{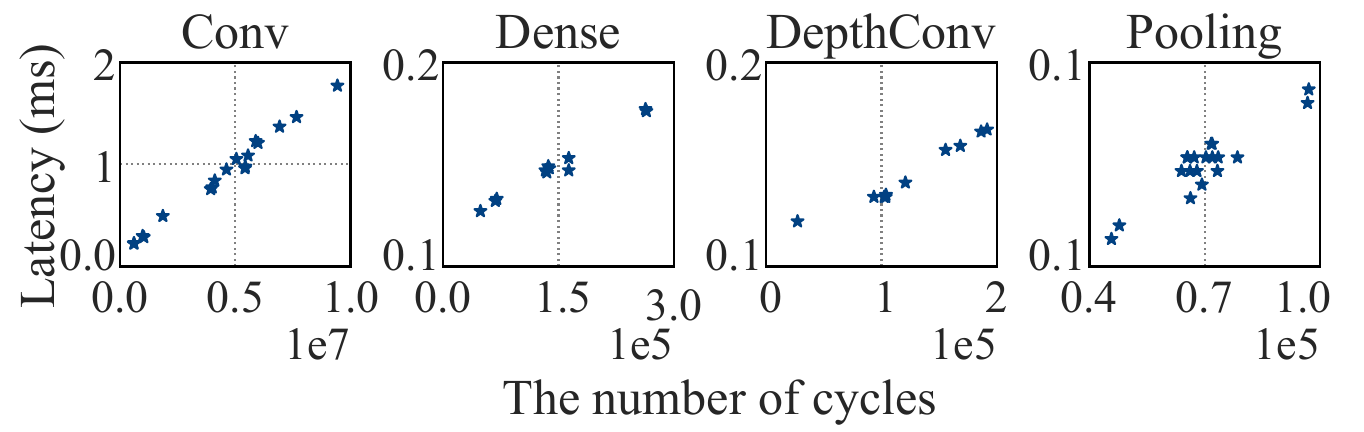}
        \vspace{-6mm}
        \caption{Our cycle-based \sysName: Latency vs number of \textit{cycles}}
        \end{subfigure}
        \vspace{-8mm}
        \caption{Linearity between our proxy and measured \textit{latency} on pure-layer models.}
        \label{fig_MAC_inst_latency}
    \vspace{-1mm}
\end{figure}
\vspace{-2mm}
\subsection{Evaluation of Proxies}
\begin{figure*}[t]
    \centering
    \begin{subfigure}[t]{0.24\textwidth}
        \centering
        \includegraphics[width=\linewidth]{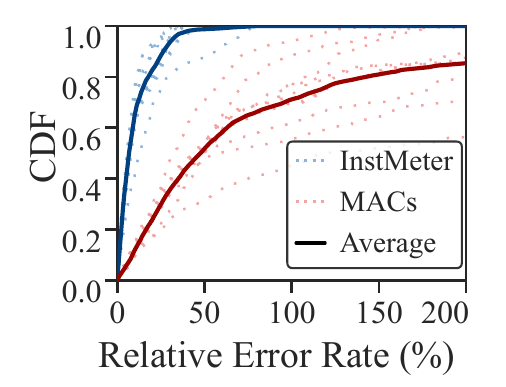}
        \vspace{-7mm}
        \caption{Ours vs. MACs (5 samples)}
        \label{fig_manual_mac_energy_datasize5}
    \end{subfigure}
    \hspace{0.5mm}
    \begin{subfigure}[t]{0.24\textwidth}
        \centering
        \includegraphics[width=\linewidth]{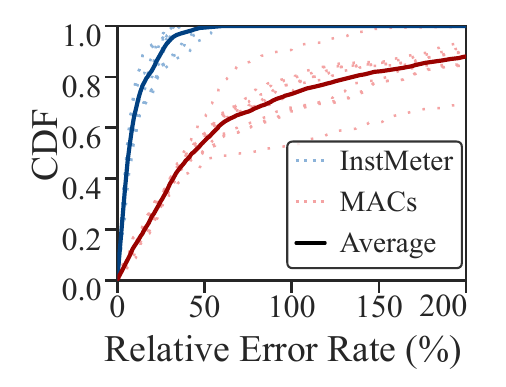}
        \vspace{-7mm}
        \caption{\blue{Ours vs. MACs (20 samples)}}
        \label{fig_manual_mac_energy_datasize20}
    \end{subfigure}
    \hspace{0.5mm}
    \begin{subfigure}[t]{0.24\textwidth}
        \centering
        \includegraphics[width=\linewidth]{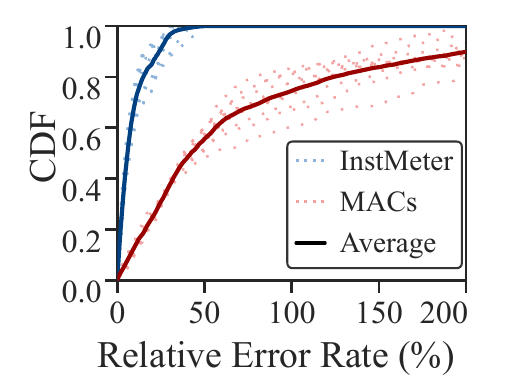}
        \vspace{-7mm}
        \caption{Ours vs. MACs (50 samples)}
        \label{fig_manual_mac_energy_datasize50}
    \end{subfigure}
    \hspace{0.5mm}
    \begin{subfigure}[t]{0.24\textwidth}
        \centering
        \includegraphics[width=\linewidth]{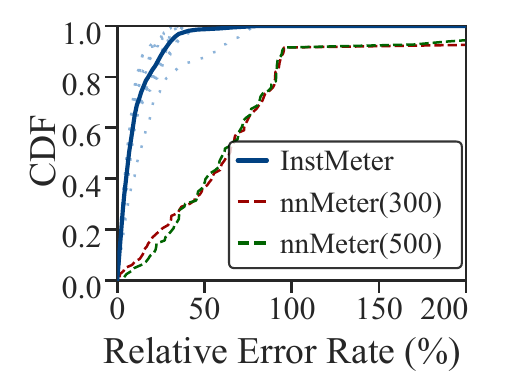}
        \vspace{-7mm}
        \caption{Ours vs. nn-Meter}
        \label{fig_manual_instmeter_nnmeter_energy}
    \end{subfigure}
    \vspace{-2mm}
    \caption{Energy prediction comparisons of our \sysName vs. MAC-based estimators and nn-Meter, with varying training data sizes on the ML-Data.}
    \label{fig_combined_energy_comparison}
\end{figure*}
We first evaluate different predictors' proxies using the single-layer dataset (SL-Data), consisting of models built from pure layers. This comparison focuses on MAC-based methods to show that, for some layers, model-level parameters have a non-linear relation with latency and energy, but our cycle-based proxies can model them \textit{linearly}.

Figure~\ref{fig_MAC_inst} and Figure~\ref{fig_MAC_inst_latency} shows the relationship between the measured energy/latency and various layer types. {We can observe that MAC-based proxies exhibit a linear relationship for \textit{dense} layers but fail to generalize across other types of layers. This occurs because MACs are not inherently tied to specific code implementations and compiler optimizations. When compiling the DL models for MCUs, TFLM and compiler optimizations introduce discrete behaviors—such as switching kernel functions based on distinct parameters—that break the linearity between MACs and energy consumption.}  
On the other hand, \sysName's cycle-based proxies maintain strong linearity across all layer types. 
Overall, the results show the linearity (superiority) of \sysName's instruction-level 
modeling for accurate energy consumption across all the DL model layers.

\vspace{-3mm}
\subsection{Comparison with SOTA}
\label{subsec_sota}

We now evaluate \sysName's performance compared to the SOTA, using the mixed-layer dataset (ML-Data).

{\textbf{\sysName vs. MACs-based estimators.}}
\blue{Figures~\ref{fig_manual_mac_energy_datasize5}-\ref{fig_manual_mac_energy_datasize50}} illustrate the relative error rates of both predictors as a function of the training data size on energy and latency predictions, respectively. 
The experiments are carried out on the mixed-layer dataset (ML-Data). Out of the 445 samples in the dataset, we use $s$ samples for training and the remaining samples for testing ($445-s$).
We set $s$ to 5, \blue{20} and 50; for each $s$, we run 10 random instances.
The results in \blue{Figures~\ref{fig_manual_mac_energy_datasize5}-\ref{fig_manual_mac_energy_datasize50}} show that \sysName achieves consistently lower error rates across all the cases with different training data sizes. The performance of the worst-case instance in \sysName is also significantly better than the performance of the best-case instance of MACs-based methods.
Specifically, if we look at the 0--90th percentiles of the averaged results (solid lines), the relative prediction error rates in \sysName remain below ~30\% across all training data sizes, whereas those of MACs-based predictors exceed 200\%, showing an improvement of over 6.5$\times$ achieved by our \sysName.

{\textbf{\sysName vs. nn-Meter.}}
With MACs-based approaches, using the same number of training samples provides a fair comparison because both proxies, MACs and cycles, use linear regression. On the other hand, nn-Meter~\cite{nnMeter} uses random forests for estimation, and thus, relying only on a few samples would lead to terrible performance. {The original nn-Meter requires thousands of training samples. While nn-Meter can learn the non-linearities with such an amount of training samples, it requires a huge amount of effort to collect such training samples for each MCU type, which is not practical due to the slow data collection on MCUs and the diversity of MCU platforms.} To create a competitive implementation of nn-Meter while showcasing the reduced training data required by \sysName, we train nn-Meter with 300 and 500 samples and \sysName with only 5 samples. 

For nn-Meter, we cannot use the ML-Data for training since it has a specific and predefined method for collecting training data. To train nn-Meter, we use their public code as is, with the only difference being that we check that the training models fit the memory requirement of MCUs. Thus, our overall training and testing phases are as follows: $(i)$ for our \sysName we use 5 training samples chosen at random from the ML-Dataset, and we repeat this process 10 times; $(ii)$ for nn-Meter, we build a new set of 500 training samples following their guidelines; and $(iii)$ to test both approaches, we use the remaining 440 samples from the ML-Dataset (after using 5 samples at random for training \sysName).

The result for \textit{energy} estimation, presented in Figure~\ref{fig_manual_instmeter_nnmeter_energy}, shows a significant gain of \sysName over nn-Meter. For all data, \sysName consistently exhibits stable and predictable error rates, ranging from 0\% to around 40\%. In contrast, nn-Meter demonstrates much higher variability, with error rates ranging over 200\%. As the amount of training data increases from 300 to 500 samples, the error rates under nn-Meter decrease and become more compact. However, the advantages of \sysName remain strong, particularly at the upper percentiles. 
At the 90th percentile, the error rates of nn-Meter reach as high as 100\%, whereas \sysName maintains its error rates within 30\%.  These results clearly show the superior stability and accuracy of \sysName.

{Overall, our results show two important points.
First,~thanks to the linearity of our approach, few samples are needed. For \sysName, there is not much improvement between using 5 and 50 samples.
Second, we attain a better performance with fewer samples. MACs-based methods use 10$\times$ more training data but provide 6.5$\times$ worse error, while nn-Meter requires 100$\times$ more training data but provides 3$\times$ worse error.}

\vspace{-2mm}
\subsection{Evaluation in NAS Scenarios}
\label{subsec_nas}

\begin{figure}[t]
    \centering
    \includegraphics[height=27mm]{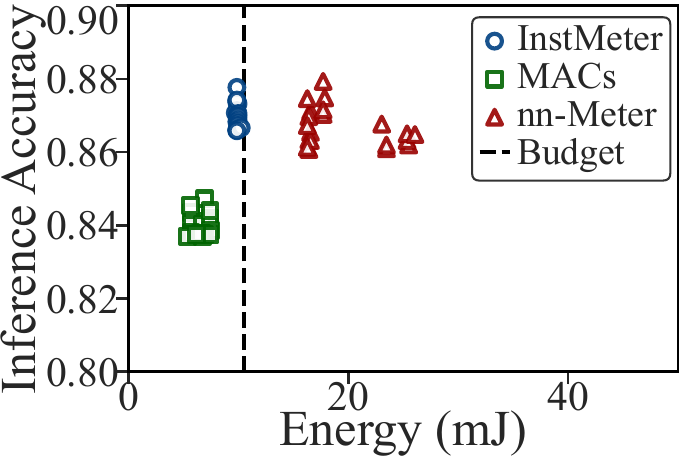}
    \hfill
    \includegraphics[height=26.5mm]{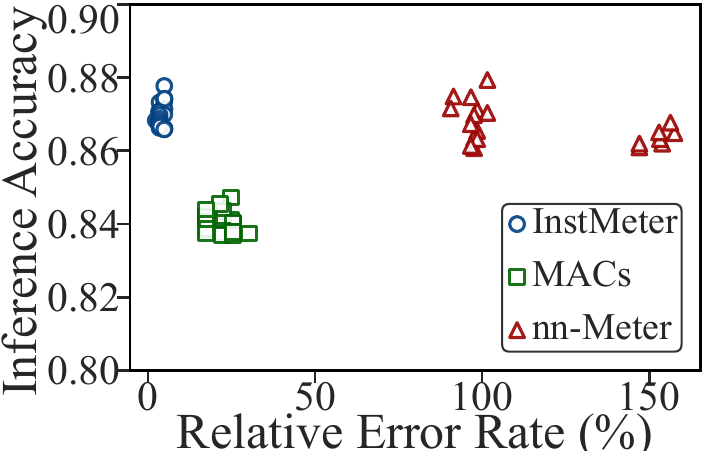}
    \vspace{-3mm}
    \caption{\blue{NAS results with 10mJ energy budget}.} 
    \label{fig_nas_e10}
    \vspace{-3mm}
\end{figure}

Until now, we have evaluated the performance of \sysName on individual models, but the key advantage of \textit{fast and accurate} predictors is on the model design.
\textbf{Neural Architecture Search (NAS)} is a family of algorithms that automatically search for optimal neural network architectures. For MCUs, predictors are central because they evaluate tens of thousands of models to maximize accuracy while satisfying constraints, such as memory, energy, or latency budgets.

In this subsection, we assess the performance of \sysName and SOTA predictors in a NAS scenario
focusing on the constraints of MCU memory and energy consumption. Accurate estimation of energy is important:
because if energy consumption is underestimated, the device will have a shorter life than expected; and if the energy consumption is overestimated, the device may sacrifice accuracy.
We use a widely adopted NAS method based on evolutionary algorithms \cite{uNAS, harvnet}. 
\blue{The NAS setup is identical across all predictors, differing only in the predictor component.}
The memory constraint corresponds to the MCU's memory size 256 KB, and the energy constraint is set to two empirical values of \blue{10~mJ} and 30~mJ. 
To compare the performance of the predictors, we report the top {20} models discovered by NAS using each predictor. \blue{We run these top models on MCUs and measure the ground-truth using our automated pipeline.}
{We configure the training sample sizes as follows: \sysName (5, ML-Data), MACs-based (50, ML-Data), nn-Meter (500, using their custom sampling method).}

\begin{figure}[t]
    \centering
    \includegraphics[height=27mm]{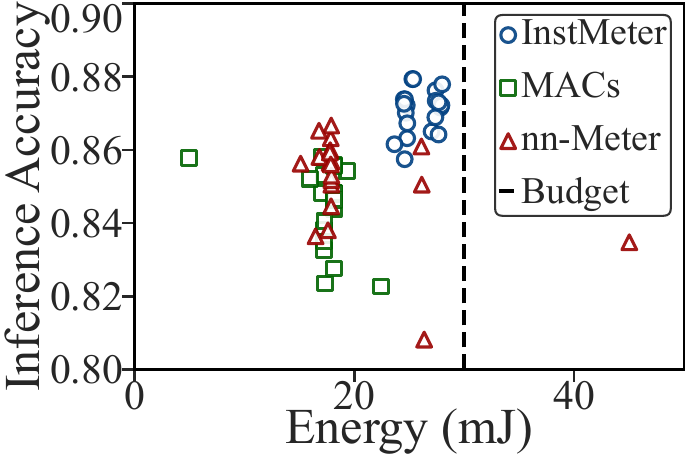}
    \hfill
    \includegraphics[height=26.5mm]{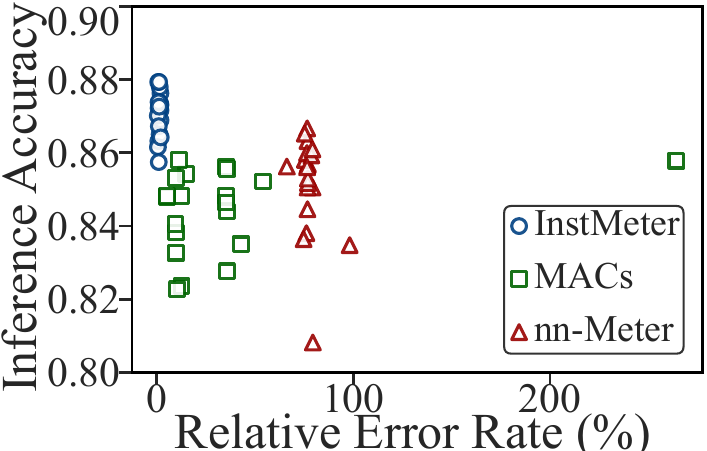}
    \vspace{-2mm}
    \caption{NAS results with 30mJ energy budget.}
    \label{fig_nas_e30}
    \vspace{-2mm}
\end{figure}

The results are presented in Figures~\ref{fig_nas_e10}-\blue{\ref{fig_nas_e30}}. 
We can observe that the models searched with the help of \sysName remain closest to the energy constraint, and have higher inference accuracy. Both MACs-based and nn-meter predictors cannot fully exploit the energy constraint, leading to models that are less accurate regarding inference. Our predictor identifies optimal models that remain closer to the provided constraints, maintaining an error rate within 10\%, while most of the other predictions exhibit errors up to 100\%. 
These results underscore the advantages of \sysName in achieving accurate predictions on models' energy consumptions even with a small training dataset. By minimizing the likelihood of predicted values exceeding or falling far below the constraint, \sysName can empower NAS to efficiently search within a valid search space, ultimately improving the efficiency and effectiveness of the NAS search process.

\vspace{-2mm}
\subsection{Generalization \& Robustness Evaluation}
\label{subsec_eval_compatibility}

We evaluate the generalization \blue{and robustness capabilities of \sysName across diverse experimental settings using the compatibility dataset. We conduct \textit{generalization experiments} (MCU architectures, TFLM versions) by individually rebuilding the instruction libraries due to the significant instruction differences. For the \textit{robustness evaluation} (compiler options, GCC versions, application scenarios, DVFS, and temperature), we reuse previously built libraries across different MCUs, explicitly evaluating \sysName's robustness without incurring frequent library regeneration.}
In each evaluation, we randomly select 5 samples for training, using the remaining samples for testing, and repeat the process 10 times.

\begin{figure}[t]
\centering
    \begin{subfigure}[t]{0.47\columnwidth}
        \centering
        \includegraphics[height=29mm]{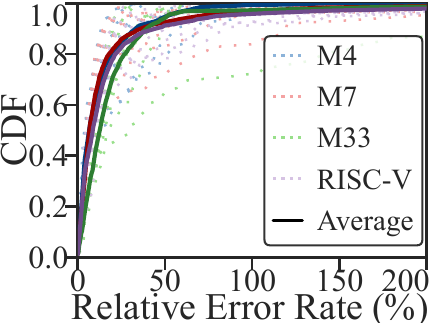}
        \vspace{-1mm}
        \caption{\blue{Different MCUs}}
        \label{fig_m7m33_energy}
    \end{subfigure}
    \hfill
    \begin{subfigure}[t]{0.51\columnwidth}
    \centering
        \includegraphics[height=29mm]{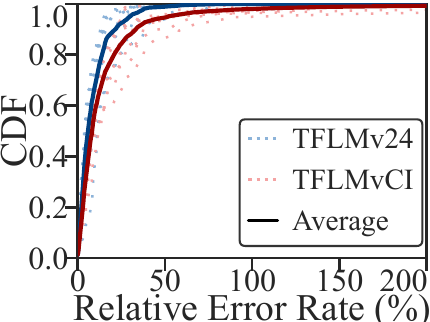}
        \vspace{-1mm}
        \caption{Different TFLM versions}
        \label{fig_TFLM24vsTFLMCI_energy}
    \end{subfigure}
    \vspace{-3mm}
    \caption{Generalization on the energy prediction across different MCUs and TFLM versions.}
    \vspace{-5mm}
\end{figure}

\begin{figure*}[h]
    \begin{subfigure}{.19\textwidth}
        \includegraphics[width=\columnwidth]{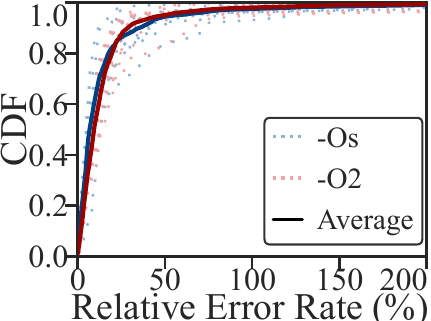}

        \caption{Compiler settings} 
        \label{fig_compiler_energy}
    \end{subfigure}
    \hfill
    \begin{subfigure}{.19\textwidth}
        \includegraphics[width=\columnwidth]{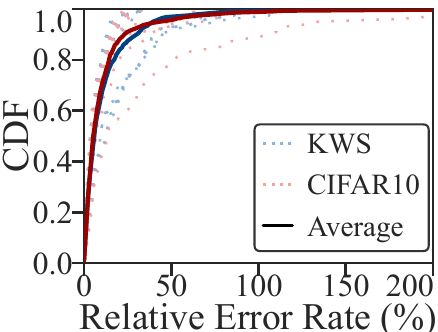}
        \caption{\blue{Applications}}
        \label{fig_app_energy}
    \end{subfigure}
    \hfill
    \begin{subfigure}{.19\textwidth}
        \includegraphics[width=\columnwidth]{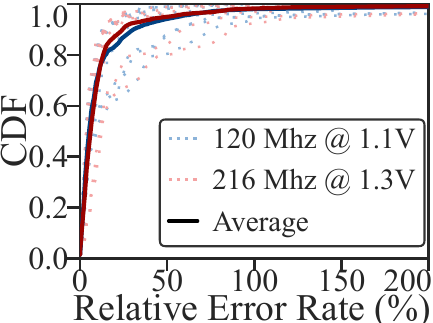}
        \caption{\blue{DVFS settings}}
        \label{fig_dvfs_energy}
    \end{subfigure}
    \hfill
    \begin{subfigure}{.19\textwidth}
        \includegraphics[width=\columnwidth]{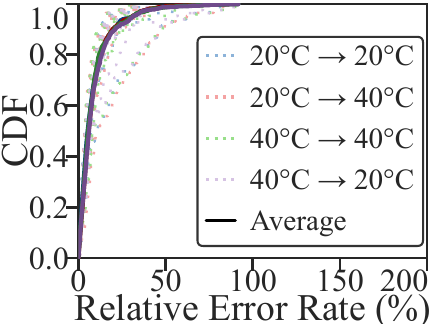}
        \caption{\blue{Temperatures}}
        \label{fig_temp_energy}
    \end{subfigure}
    \hfill
    \begin{subfigure}{.19\textwidth}
        \includegraphics[width=\columnwidth]{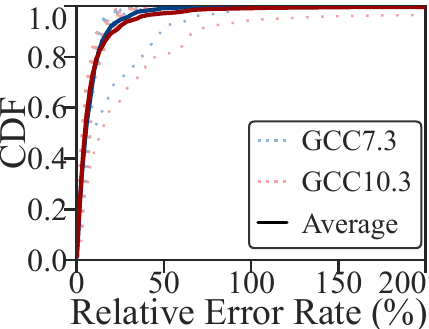}
        \caption{\blue{GCC versions}}
        \label{fig_gcc_energy}
    \end{subfigure}
    \vspace{-2mm}
    \caption{Robustness of energy prediction across different TFLM versions, applications, DVFS settings, temperatures, and GCC compiler versions.}
    \vspace{-3mm}
\end{figure*}

{\textbf{Generalization on other ARM MCUs and the RISC-V-based MCU.}}
In prior experiments, the Cortex-M4 platform is used as the default for evaluation. Here, we further apply \sysName to another three MCUs: Cortex-M7, M33 and \blue{ {RISC-V-based ESP32-C3}}. To extend our framework, we follow the procedure described in Section~\ref{sec_method}. Specifically, we compile model functions to generate \blue{disasm code tailored for new MCUs, and replace the Cortex-M4 files (source and disasm files) with new MCUs' counterparts.}  
We then rerun our framework to establish mappings and perform the corresponding operator-to-instructions profiling. The generated dictionaries are used for modeling on the corresponding platforms. 
The results are summarized in Figure~\ref{fig_m7m33_energy}. Overall, our \sysName demonstrates stable performance across the M4, M7, M33 and \blue{ {RISC-V-based ESP32-C3}} platforms. The average relative error almost remains below $50$\% across all the platforms. For 95\% of the models, Cortex-M7's, M4's, and \blue{ {ESP32-C3's}} error rates are consistently below $30$\%, while for Cortex-M33, our \sysName remains stable below $40$\%. These results confirm the robust cross-platform compatibility of \sysName and its ability to maintain high prediction accuracy across diverse MCU platforms. The slightly higher error rate in the M33 platform is mainly due to the lack of detailed cycle counts per instruction in its manual, so we need to use the corresponding details in M4 to build the predictor.
\blue{The instruction library overhead is quantified as follows: storage sizes (31-32~KB for the three Cortex-M MCUs, and 58 KB for the RISC-V MCU), and query latencies (84 ms/query for the Cortex-M MCUs, and 154 ms/query for the RISC-V MCU), averaged over 100 NAS-generated models.}

{\textbf{Generalization across different TFLM versions.}}
Different versions of the TFLM library implement the model operators with varying code optimizations,
potentially affecting the prediction accuracy of energy consumption. To evaluate the compatibility and robustness of our predictor across various TFLM versions, we conduct experiments using two widely adopted versions: TFLMv2.4 and TFLMvCI. 
\blue{We generate version-specific instruction libraries by directly utilizing the source files from each TFLM version, precisely modeling their implementation differences.}
The evaluation results are shown in Figure~\ref{fig_TFLM24vsTFLMCI_energy}. We can observe strong compatibility of \sysName across both TFLM versions. For 90\% of the models, the relative error rates for the predicted energy cost remain below $40$\%, which validates \sysName's consistent performance. TFLMvCI has slightly higher relative error rates compared to TFLMv2.4. Nevertheless, the overall results demonstrate \sysName's robustness in accommodating variations across software versions.

{\textbf{Robustness with different compiler settings.}}
The impact of compiler options on the prediction of energy consumption is often overlooked in prior studies. However, compiler options can significantly influence the execution order and frequency of program instructions, thereby affecting the overall energy cost. To investigate these effects, we evaluate two representative compiler options: \texttt{-Os} (optimized for minimal code size) and \texttt{-O2} (optimized for maximum execution speed). 
\blue{We leverage the profiling libraries compiled at \texttt{-O0} to directly predict performance at \texttt{-O2} and \texttt{-Os}.}
The results are shown in Figure~\ref{fig_compiler_energy}. We can observe the consistent performance of \sysName across both compiler settings. For 90\% of the models, the relative error rates for both \texttt{-Os} and \texttt{-O2} remain within $30$\%. \blue{The \texttt{-O2} option yields slightly less accurate results, because we use libraries profiled at \texttt{-O0}, while the option \texttt{-O2} introduces structural changes to the code, such as loop unrolling and other optimizations for improving the execution speed, which slightly degrade prediction accuracy.} In summary, the results in Figure~\ref{fig_compiler_energy} demonstrate that our \sysName maintains strong adaptability and robustness across different compilation environments.

{{\textbf{Robustness across different applications.}}
{Besides the default KWS application, we also evaluate \sysName for another representative application: \textit{image recognition} (using the CIFAR-10 dataset). KWS and image recognition represent distinct tasks with different data types, model architectures, and operator patterns. The DL models for KWS are lightweight and process short audio clips, whereas the DL models for the image recognition application are deeper and take images as input. The results are shown in Figure~\ref{fig_app_energy}. We can observe that \sysName maintains strong prediction accuracy across both tasks, with over 90\% of the models having a relative error of less than 20\%. This confirms \sysName's robustness across different DL tasks.}}

{\textbf{Robustness under DVFS configurations.}}  
\blue{We further evaluate \sysName under two DVFS configurations on the STM32F767ZI MCU to examine its robustness under realistic power-performance trade-offs. Specifically, we measure the ground-truth energy and latency using the OTII ACE PRO power meter at a high-performance setting (216MHz@1.3V) and a low-power setting (120MHz@1.1V). Note that, following common practice~\cite{pros, eenet}, the DVFS setting is configured at the beginning of inference execution and remains fixed throughout inference, as intra-inference voltage/frequency switching often introduces significant overhead. Figure~\ref{fig_dvfs_energy} shows, across these two DVFS operating points, \sysName achieves a relative prediction error of less than 25\% for over 90\% of the models, confirming its reliable prediction accuracy and adaptability to different DVFS settings.}

\textbf{Robustness under different temperature conditions.}
\blue{
We also evaluate the robustness of \textit{InstMeter}'s predictions under
different temperatures.
We use an FLIR E4 infrared camera to monitor the MCU chip's surface temperature. Specifically, we measure the ground-truth energy cost at two distinct temperature conditions: an indoor scenario with the chip's surface temperature staying about $21\pm2^\circ$C, and an outdoor scenario with the chip's surface temperature reaching $43\pm2^\circ$C. The performance evaluation results of \sysName are shown in Figure~\ref{fig_temp_energy}. We can observe that \sysName, trained using 5 samples at 21\textdegree C, can accurately predict the energy cost at both 21\textdegree C and 43\textdegree C, with nearly identical accuracy. Likewise, \sysName trained at 43\textdegree C maintains similar prediction accuracy across both temperature conditions. This cross-temperature evaluation demonstrates that \textit{InstMeter}'s prediction error remains consistently below around 22\% for over 90\% of the tested models, confirming its stability and reliability across typical operating temperature ranges.}

\textbf{Robustness across different GCC versions.}
\blue{Finally,~we evaluate \sysName across two widely-used GCC compiler versions (v7.3, v10.3) to assess its robustness against compiler-induced variations. Different compiler versions can impact instruction scheduling, optimization behaviors, and resulting execution patterns, thereby affecting the energy cost. The evaluation results are shown in  Figure~\ref{fig_gcc_energy}. We can see that \sysName achieves consistently accurate predictions across both GCC versions, with over 90\% of the models exhibiting a relative error below 20\%. This shows \sysName's strong resilience to variations caused by different compiler versions.}

\vspace{-4mm}
\section{Discussion \& Related Work}
\label{sec_relatedwork}

{\textbf{Generalization to MCUs with hardware accelerators.}
{Our current implementation of \sysName relies on the TensorFlow Lite for Microcontrollers (TFLM) framework, which does not directly support hardware accelerators (e.g., Arduino Nicola Voice, MAX78000) due to their different software stacks. These accelerators use their own closed-source tools instead of TFLM, making the generalization of  \sysName on them challenging. We leave extending TFLM to directly support such accelerators as a future research direction.}}

{\textbf{Impact of thermal physics.} Extreme temperature affects physical properties like leakage current and MOSFET thresholds. These thermal variances are treated as environmental factors beyond the scope of our static analysis. Consequently, the dynamic performance impact triggered by thermal throttling is not currently modeled in our system. We leave extending our framework to account for these temperature-dependent fluctuations as a future work.}

\textbf{Limitation of the loop-mapping algorithm.}
\blue{Our predictor has higher errors at the aggressive compiler optimization levels such as \texttt{-O2}. This is because aggressive optimizations greatly alter loop structures through transformations like loop fusion or fission, making structural mapping more challenging. We leave 
this improvement as future work.}

\textbf{Requiring access to source and {disasm} codes.}
\blue{\sysName requires access to both source and {disasm} code, which may pose technical challenges for adapting to new MCU architectures or compiler versions when help is needed. To mitigate this, we document our step-by-step methods, openly provide the specific TFLM version used, and release our prebuilt instruction library. These resources aim to significantly simplify the customized adaptations for future researchers.
}

\textbf{Energy and latency prediction in NAS.}
Studies~\cite{monas, Auritus} are based on direct device measurements, which are accurate but incur significant overhead. To improve efficiency, some solutions have adopted lookup tables~\cite{proxylessnas} or MACs-based models~\cite{micronets, uNAS, harvnet} to balance prediction speed and accuracy. Recent advances have leveraged non-linear models, such as BRP-NAS \cite{brp-nas}, which employs graph convolutional networks for structural learning, and nn-Meter \cite{nnMeter}, which uses layer fusion rules and random forests for greater accuracy. However, they need extensive training data (1000's of samples). Our \sysName leverages hardware-specific instructions, the lowest-level features of MCUs, to accurately estimate the energy and latency of deep learning models. Due to its strong linearity property, \sysName is simpler and more accurate.
\vspace{-2mm}
\section{Conclusion}
\label{sec_conclusion}

We designed \sysName, an instruction-level framework that can rapidly construct linear predictors to accurately estimate the energy and latency costs of deep learning model inference on MCUs. \sysName achieves superior performance compared to the state-of-the-art methods. Extensive experiments demonstrated the generalization and robustness performance of the proposed \sysName across various \blue{MCUs (ARM Cortex-M4, M7, M33, RISC-V-based ESP32C3), GCC (v7.3, v10.3), applications (keyword spotting, image recognition), DVFS, temperatures (21\textdegree{}C, 43\textdegree{}C)}, compiler optimizations (-Os, -O2), and TFLM versions (v2.4, \blue{vCI}).

\vspace{5mm}
\noindent \textbf{Acknowledgment:} This work is partly supported by the NWO \textit{LuxSenz} project funded through the TOP Grant (Module 1) under project number 612.001.854, and the EU's Horizon Europe \textit{HarmonicAI} project under the HORIZON-MSCA-2022-SE-01 scheme with grant agreement number 101131117. 

\normalsize

\bibliographystyle{ACM-Reference-Format}
\bibliography{ref}

\begin{appendices}

\section{Further Discussions and Related Work}

\textbf{Extension to CPUs and GPUs.}
\blue{Extending \sysName to CPUs and GPUs is challenging because their execution is far less deterministic than on MCUs. CPUs introduce deep pipelines and caches, while GPUs add massive parallelism, thread divergence, and memory coalescing. Moreover, heterogeneous CPU–GPU coordination and data movement introduce further variability in the latency and energy cost of deep-learning model inference. Modeling these effects will require detailed micro-architectural knowledge, but is an important direction for follow-up research of \sysName.}

\setcounter{figure}{16}

\begin{figure}[b!]
\vspace{-2mm}
    \centering
    \begin{subfigure}[t]{0.235\textwidth}
        \centering
        \includegraphics[width=\linewidth]{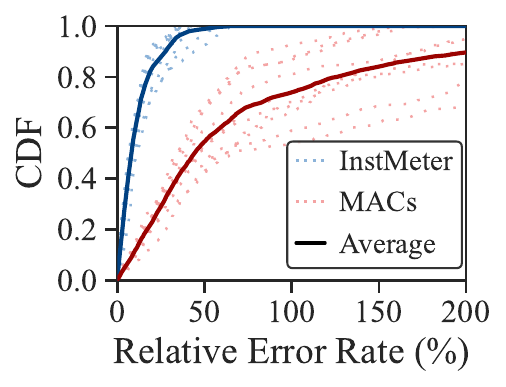}
        \vspace{-6mm}
        \caption{Ours vs. MACs (5 samples)}
        \label{fig_manual_mac_latency_datasize5}
    \end{subfigure}
    \hfill
    \begin{subfigure}[t]{0.235\textwidth}
        \centering
        \includegraphics[width=\linewidth]{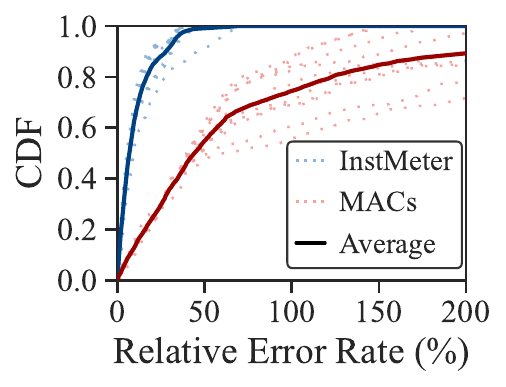}
        \vspace{-6mm}
        \caption{\blue{Ours vs MACs (20 samples)}}
    \label{fig_manual_mac_latency_datasize20}
    \end{subfigure}
    \\
    \begin{subfigure}[t]{0.235\textwidth}
        \centering
        \includegraphics[width=\linewidth]{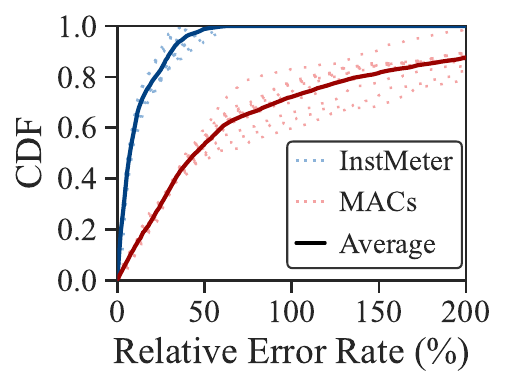}
        \vspace{-6mm}
        \caption{Ours vs. MACs (50 samples)}
        \label{fig_manual_mac_latency_datasize50}
    \end{subfigure}
    \hfill
    \begin{subfigure}[t]{0.235\textwidth}
        \centering
        \includegraphics[width=\linewidth]{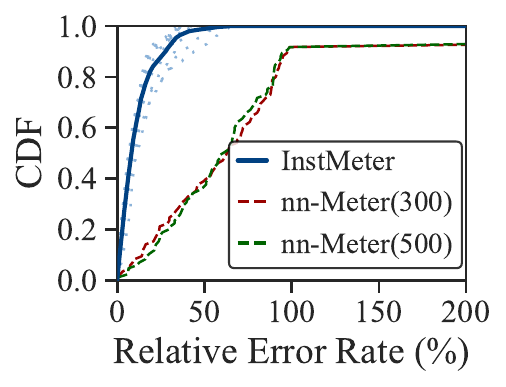}
        \vspace{-6mm}
        \caption{Ours vs. nn-Meter}
        \label{fig_manual_instmeter_nnmeter_latency}
    \end{subfigure}
    \vspace{-3mm}
    \caption{The \textit{latency} prediction comparisons of our \sysName vs. MAC-based estimators and nn-Meter, with varying training data sizes on the ML-Data.}
    \label{fig_combined_energy_comparison}
\end{figure}



\textbf{Source-{disasm} code matching.}
BinPro uses the string literals and function invocation information of the target codes, and then employs the Hungarian algorithm to match these two languages \cite{binpro}. Building on this, B2SFinder enhances the feature set by incorporating `if' and `switch' statements, paired with a weighted matching algorithm for improved accuracy \cite{b2sfinder}. Further advancements are seen with CodeCMR, which utilizes a graph neural network to analyze control flow graphs and an LSTM model to process traditional literals. These are then linked through dense layers to facilitate effective matching \cite{codecmr}. Expanding on cross-language capabilities, tools like study \cite{cross} and BinaryAI \cite{binaryai} convert \blue{disasm} code into an intermediate representation (IR) and pseudocode, respectively. These are then processed using transformer-based DL algorithms. All these methods focus on library and function levels matching. In \sysName, our design centers on loop-level code matching, providing a more granulated and precise analysis. In addition, CodeCMR, BinaryAI and~\cite{cross} are based on DL-related methods, which require extensive data collection. Building upon BinPro~\cite{binpro} and B2SFinder~\cite{b2sfinder}, we extract structural and operator information from source and \blue{disasm} code, allowing us to achieve finer-granularity loop-level source-\blue{disasm} code matching. 

\section{Evaluation on the Latency Prediction}
\label{sec_eva_latency}

\subsection{Latency: Comparison with SOTA}
\label{subsec_latency_macsVSinstmeter}

\textbf{\sysName vs. MACs-based estimators.}
Figures~\ref{fig_manual_mac_latency_datasize5}-\ref{fig_manual_mac_latency_datasize50} shows the relative latency prediction errors of \sysName compared to MACs-based predictors, using the mixed-layer dataset (ML-Data) with training data sizes of 5, 20 and 50 samples, each tested over 10 random trials.
\sysName consistently achieves significantly lower prediction errors across all training set sizes and trials. Specifically, the 90th percentile relative error for \sysName remains below approximately 30\%, whereas MACs-based methods exceed 200\%, demonstrating over 6.5$\times$ improvement. These results clearly confirm the superior robustness and accuracy of our cycle-based latency prediction compared to MAC-based estimators.

\begin{figure}[b!]
\vspace{-2mm}
    \begin{subfigure}[t]{0.47\columnwidth}
        \centering
        \includegraphics[height=28mm]{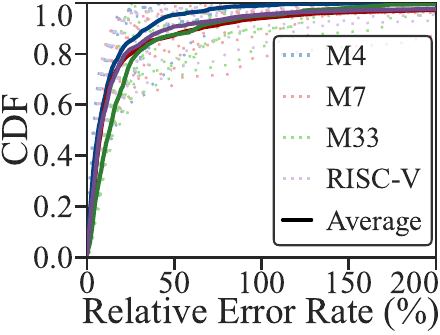}
        \caption{\blue{Different MCUs}}
        \label{fig_m7m33_latency}
    \end{subfigure}
    \hfill
    \begin{subfigure}[t]{0.51\columnwidth}
    \centering
     \includegraphics[height=28mm]{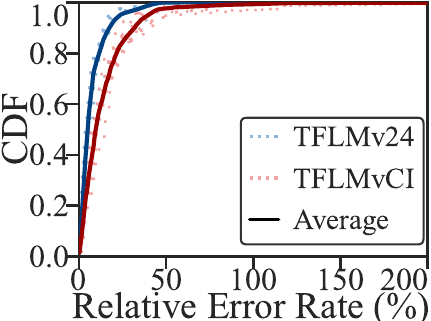}
        \caption{Different TFLM versions}
        \label{fig_TFLM24vsTFLMCI_latency}
    \end{subfigure}
    \vspace{-3mm}
    \caption{Generalization on \textit{latency} prediction across different MCUs and TFLM versions.}
\end{figure}


\begin{figure*}[t]
    \begin{subfigure}{.19\textwidth}
        \includegraphics[width=\columnwidth]{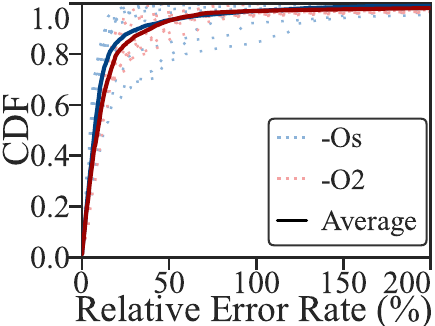}
        \caption{Compiler settings} 
        \label{fig_compiler_latency}
    \end{subfigure}
    \hfill
    \begin{subfigure}{.19\textwidth}
        \includegraphics[width=\columnwidth]{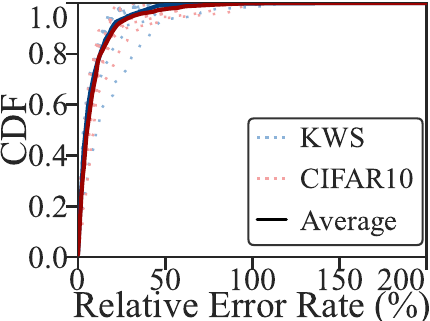}
        \caption{\blue{Applications}}
        \label{fig_app_latency}
    \end{subfigure}
    \hfill
    \begin{subfigure}{.19\textwidth}
        \includegraphics[width=\columnwidth]{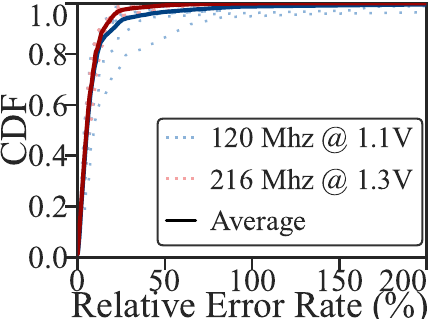}
        \caption{\blue{DVFS settings}}
        \label{fig_dvfs_latency}
    \end{subfigure}
    \hfill
    \begin{subfigure}{.19\textwidth}
        \includegraphics[width=\columnwidth]{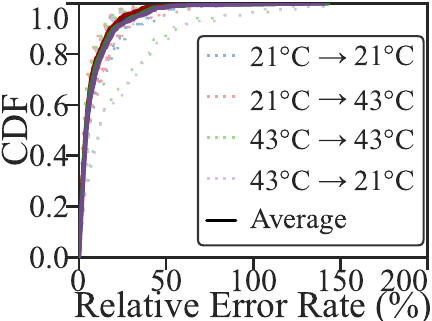}
        \caption{\blue{Temperatures}}
        \label{fig_temp_latency}
    \end{subfigure}
    \hfill
    \begin{subfigure}{.19\textwidth}
        \includegraphics[width=\columnwidth]{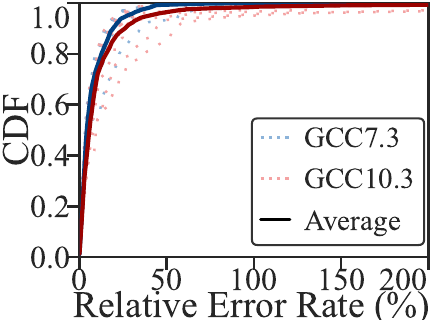}
        \caption{\blue{GCC versions}}
        \label{fig_gcc_latency}
    \end{subfigure}
    \vspace{-1mm}
    \caption{Robustness on the \textit{latency} prediction across different TFLM versions, applications, DVFS settings, temperatures, and GCC compilers.}
\end{figure*}

\textbf{\sysName vs. nn-Meter.}
Figure~\ref{fig_manual_instmeter_nnmeter_latency} compares latency estimation performance between \sysName and nn-Meter~\cite{nnMeter}, where nn-Meter is trained with 300 and 500 samples using its original sampling approach, and \sysName uses only 5 samples from ML-Data. Results show that \sysName achieves significantly lower and more stable error rates (0\%–40\%) compared to nn-Meter (exceeding 200\%). Specifically, at the 90th percentile, nn-Meter's errors reach about 100\%, whereas \sysName consistently maintains error rates below 30\%, demonstrating over 3$\times$ improvement with 100$\times$ fewer samples.



\vspace{-2mm}
\subsection{Latency: Generalization \& Robustness}

\textbf{Generalization on other ARM MCUs and the RISC-V-based MCU.} 
We evaluate the cross-platform generalization of \sysName's latency prediction from Cortex-M4 to additional MCUs: Cortex-M7, M33, and RISC-V-based ESP32-C3. For each MCU, we individually rebuild instruction libraries using platform-specific disassembled codes. As shown in Figure~\ref{fig_m7m33_latency}, \sysName achieves consistent accuracy across these diverse architectures, with latency prediction errors under 30\% for M4, M7, and ESP32-C3, and under 40\% for M33. The slightly higher error on M33 is due to approximating CPI values from M4, caused by incomplete official documentation. These results confirm that \sysName reliably generalizes latency estimation across diverse MCU platforms.

\textbf{Generalization across different TFLM versions.}
We further evaluate \sysName's latency prediction generalization across two popular TFLM library versions: v2.4 and vCI. For each version, we generate instruction libraries using corresponding source files, capturing their implementation differences explicitly. The latency prediction results, shown in Figure~\ref{fig_TFLM24vsTFLMCI_latency}, demonstrate strong generalization, with errors consistently below 40\% for 90\% of the models. While TFLMvCI yields slightly higher errors than v2.4, these results confirm {\sysName's} capability to generalize latency prediction across software library variations.

\textbf{Robustness with different compiler settings.}
We assess the robustness of \sysName's latency predictions under two widely used compiler optimizations: \texttt{-Os} (size optimization) and \texttt{-O2} (speed optimization). Predictions are made by reusing the Cortex-M7 instruction library compiled at \texttt{-O0}, deliberately introducing instruction-level variations. The results, shown in Figure~\ref{fig_compiler_latency}, indicate robust prediction accuracy, with latency errors within 30\% for over 90\% of the models. Higher errors with \texttt{-O2} are attributable to aggressive code optimizations, such as loop unrolling, which slightly degrade prediction accuracy. These results confirm the resilience of \sysName's latency prediction against common compiler variations.

\textbf{Robustness across different applications.}
\blue{We further test the latency prediction performance of \sysName across the keyword spotting (KWS) and image recognition (using CIFAR-10 dataset) applications. We reuse the instruction library built from Cortex-M4 (\texttt{-O0}), despite variations in data types, model complexity, and operator patterns between tasks. The latency prediction results are given in Figure~\ref{fig_app_latency}, showing that \sysName can consistently achieve errors below 25\% for over 90\% of models across the two applications. This demonstrates \sysName's reliable latency prediction across diverse DL workloads and input modalities.}

\textbf{Robustness under DVFS configurations.}
\blue{We evaluate the robustness of \sysName's latency predictions across realistic DVFS settings (216MHz@1.3V and 120MHz@1.1V) on the STM32F767ZI MCU, reusing the instruction library generated from Teensy 4.1 (Cortex-M7, \texttt{-O0}). Ground-truth latency measurements are acquired using accurate instrumentation. The results, as shown in Figure~\ref{fig_dvfs_latency}, indicate that over 90\% of the latency predictions have errors under 20\% across both DVFS points. These results confirm that \sysName reliably accommodates latency variations induced by typical DVFS settings encountered in embedded scenarios.}


\textbf{Robustness under different temperature conditions.}
\blue{We further evaluate \sysName's latency prediction stability under realistic temperature variations, predicting latency at MCU chip temperatures of 21°C and 43°C using a common instruction library (\texttt{-O0}). The chip temperatures are measured with an FLIR E4 infrared camera. The results, shown in Figure~\ref{fig_temp_latency}, demonstrate consistently accurate latency predictions, with over 90\% of models showing errors below 25\% at both temperature conditions. These results highlight the minimal impact of temperature variations on \sysName's latency prediction, demonstrating practical robustness in typical embedded operating environments.}

\textbf{Robustness across different GCC versions.}
\blue{Lastly, we evaluate the latency prediction of \sysName across GCC compiler versions (v7.3 and v10.3).
Different compiler versions influence instruction scheduling and optimization, potentially affecting latency. Despite such potential variations, the results, as shown in Figure~\ref{fig_gcc_latency}, demonstrate that \sysName can achieve latency prediction errors below 20\% for over 90\% of tested models. This result emphasizes \sysName's strong resilience to compiler-induced variations, affirming its practical robustness in diverse compilation environments.}

\end{appendices}

\end{document}